\documentclass[10pt,twocolumn,letterpaper]{article}

\usepackage[pagenumbers]{cvpr} 
\usepackage[accsupp]{axessibility}  

\usepackage{amsfonts}       
\usepackage{nicefrac}       

\usepackage{graphicx}
\usepackage{comment}
\usepackage{amsmath,amssymb,amstext} 

\usepackage{gensymb}

\usepackage{relsize}
\usepackage[T1]{fontenc}
\usepackage[scaled=0.90]{inconsolata}
\usepackage[latin1]{inputenc}
\usepackage[english]{babel}

\usepackage{epsfig}
\usepackage{graphicx}
\usepackage{wrapfig}
\usepackage[belowskip=2pt,labelfont=bf,aboveskip=0pt,font=small,labelsep=period]{caption}
\usepackage{array}

\usepackage[belowskip=0pt,aboveskip=0pt,font=small]{subcaption}
\usepackage{longtable}
\usepackage{microtype}

\usepackage{textcomp}
\usepackage{stmaryrd}
\SetSymbolFont{stmry}{bold}{U}{stmry}{m}{n}
\usepackage{upgreek}
\usepackage{bm}
\usepackage{cases}
\usepackage{mathtools}
\usepackage{arydshln}
\usepackage{multirow}

\usepackage{color}
\usepackage[dvipsnames]{xcolor}
\definecolor{JalapenoRed}{RGB}{183,21,64}
\definecolor{Belize}{RGB}{41,128,185}
\definecolor{Amour}{RGB}{238,82,83}
\usepackage{colortbl}

\usepackage[pagebackref=true,breaklinks=true,colorlinks,bookmarks=false,allcolors=JalapenoRed]{hyperref}

\usepackage[capitalize]{cleveref}
\crefname{section}{Sec.}{Secs.}
\Crefname{section}{Section}{Sections}
\Crefname{table}{Table}{Tables}
\crefname{table}{Tab.}{Tabs.}

\usepackage{algorithm}
\usepackage{algpseudocode}

\usepackage{multirow}
\usepackage{rotating}
\usepackage{booktabs}
\usepackage{etoolbox}
\newcounter{magicrownumbers}
\preto\tabular{\setcounter{magicrownumbers}{0}}
\newcommand\rownumber{\stepcounter{magicrownumbers}\arabic{magicrownumbers})\,}

\usepackage{enumitem}
\usepackage[olditem,oldenum]{paralist}

\usepackage{alltt}
\usepackage{listings}

\usepackage{url}
\usepackage{xspace}
\usepackage{comment}
\usepackage{cuted}
\usepackage{caption}

\usepackage{quoting}
\quotingsetup{vskip=0pt}
\usepackage{epigraph}
\usepackage[normalem]{ulem}

\setdefaultleftmargin{.2cm}{.2cm}{}{}{}{}
\setlist{leftmargin=.2cm}

\usepackage[textsize=tiny]{todonotes}

\newcommand{\SP}{\text{SP}\xspace}
\newcommand{\ESP}{\text{\texttt{e}SP}\xspace}
\renewcommand{\hat}[1]{\widehat{#1}}
\newcommand{\bcdot}{\boldsymbol{\cdot}}

\usepackage{mysymbols}

\graphicspath{{images/}}

\usepackage{graphicx}
\usepackage{amsmath}
\usepackage{amssymb}
\usepackage{booktabs}
\usepackage{tablefootnote}
\usepackage{float}
\usepackage{caption}

\newcolumntype{=}{
  >{\gdef\@rowstyle{}}%
}

\newcolumntype{+}{
  >{\@rowstyle}%
}

\usepackage[capitalize]{cleveref}
\crefname{section}{Sec.}{Secs.}
\Crefname{section}{Section}{Sections}
\Crefname{table}{Table}{Tables}
\crefname{table}{Tab.}{Tabs.}

\def\cvprPaperID{1504} 
\def\confName{CVPR}
\def\confYear{2024}

\begin{document}

\title{Seeing the Unseen: Visual Common Sense for Semantic Placement}

\author{
  Ram Ramrakhya$^{1*}$ 
  \,\,
  Aniruddha Kembhavi$^{2}$ \,\,
  Dhruv Batra$^{1}$ \,\,
  Zsolt Kira$^{1}$ \,\, 
  Kuo-Hao Zeng$^{2\dagger}$
  \,\,
  Luca Weihs$^{2\dagger}$
  \,\,
  \\[0.3em]
  {\normalsize $^1$Georgia Institute of Technology \quad $^2$PRIOR @ Allen Institute of AI}\\[-0.1em]
  {\tt\small $^1$\{ram.ramrakhya,dbatra,zk15\}@gatech.edu}
  \quad {\tt\small $^2$\{anik,khzeng,lucaw\}@allenai.org} \\[-0.1em]
  {\tt\small Project Page: \href{https://ram81.github.io/projects/seeing-unseen.html}{\tt{ram81.github.io/projects/seeing-unseen}}}
}

\let\svthefootnote\thefootnote
\newcommand\freefootnote[1]{%
  \let\thefootnote\relax%
  \footnotetext{#1}%
  \let\thefootnote\svthefootnote%
}

\twocolumn[{
\renewcommand\twocolumn[1][]{#1}
\maketitle
\vspace*{-0.25in}
\centering
\captionsetup{type=figure}\includegraphics[width=0.95\linewidth]{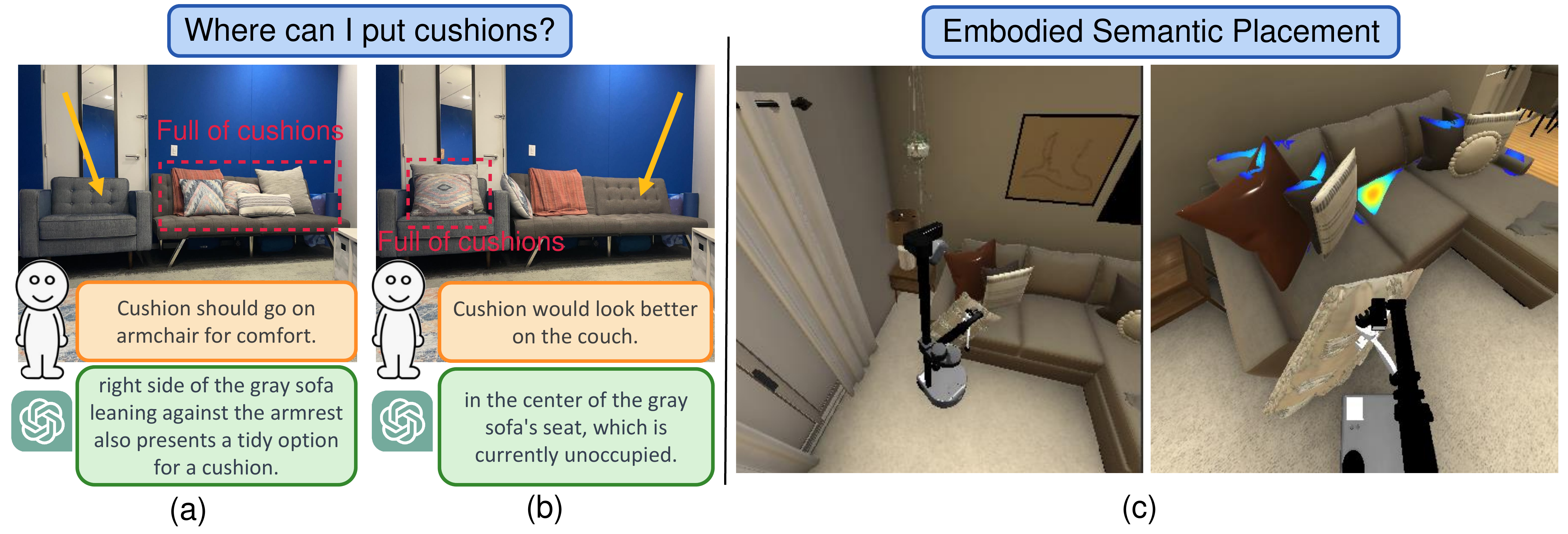}
\captionof{figure}{\textbf{Semantic Placement}. Consider asking an agent to place cushions in a living room. In (a), the couch on the right is already full with cushions, and a natural human preference would be to place the cushion against the backrest of the armchair. In (b), a natural placement preference would be center of the couch. We propose the problem of Semantic Placement (SP) --  given an image and a name of an object, a vision system must predict a semantic mask indicating a valid placement for the object in the image. For both (a) and (b) GPT4V gives meaningful natural language responses but, as we show, struggles to localize regions precisely in pixel space.
(c) Our SP predictions enable a Stretch robot~\cite{Kemp2022StretchRobot} from Hello Robot to perform Embodied Semantic Placement (\ESP) task within a photorealistic simulated environment.\\[0.15in]}
\label{fig:task}
\vspace{-1em}
}]

\freefootnote{$^{*}$Work done as part of the internship at PRIOR @ AI2}
\freefootnote{$^{\dagger}$Equal advising}


\begin{abstract}
\vspace{-1em}
Computer vision tasks typically involve describing what is present in an image (\eg classification,  detection, segmentation, and captioning). 
We study a visual common sense task that requires understanding `what is not present'.
Specifically, given an image (\eg of a living room) and a name of an object ("cushion"), a vision system is asked to predict semantically-meaningful regions (masks or bounding boxes) in the image where that object \emph{could be placed} or is likely be placed by humans (\eg on the sofa). 
We call this task: Semantic Placement (SP) and believe that such common-sense visual understanding is critical for assitive robots (tidying a house), AR devices (automatically rendering an object in the user's space), and visually-grounded chatbots with common sense. 
Studying the invisible is hard. Datasets for image description are typically constructed by curating relevant images (e.g. via image search with object names) and asking humans to annotate the contents of the image; neither of those two steps are straightforward for objects \emph{not} present in the image. 
%
We overcome this challenge by operating in the opposite direction: we start with an image of an object in context, which is easy to find online, and then \emph{remove} that object from the image via inpainting.
This automated pipeline converts unstructured web data into a dataset comprising pairs of images with/without the object.
With this proposed data generation pipeline, we collect a novel dataset, containing ${\sim}1.3$M images across $9$ object categories.
We then train a \SP prediction model, called CLIP-UNet, on our dataset. The CLIP-UNet outperforms existing VLMs and baselines that combine semantic priors with object detectors,
generalizes well to real-world and simulated images and exhibits semantics-aware reasoning for object placement.
In our user studies, we find that the \SP masks predicted by CLIP-UNet are favored $43.7\%$ and $31.3\%$ times when comparing against the $4$ SP baselines on real and simulated images.
In addition, leveraging \SP mask predictions from CLIP-UNet enables downstream applications like building tidying robots in indoor environments.

\end{abstract}

\vspace{-1.5em}
\section{Introduction}
\label{sec:intro}

%
When tasked with putting away a cushion in a home, humans quickly bring to bear extensive priors about how cushions are used and where they are most frequently placed. For instance, cushions are generally put on or near seating areas (\eg, on a couch). 
%
However, these priors themselves are not enough: consider an example living room shown in~\cref{fig:task}(a). As shown in the figure, the couch already has cushions on both armrests so, to avoid redundancy, one might place the cushion against the back of the armchair for the comfort of anyone who might later sit upon it.
On the other hand, given the same task with the image from~\cref{fig:task}(b), the answer might change to placing the cushion in center of the couch to give the room a more aesthetically pleasant feel as the armchair already has a cushion on it.
Notice that the answer from humans about object placement changes based on changes in the visual context.
%
%
We call this task Semantic Placement (SP), and believe that such common-sense visual understanding is critical for assistive robots (tidying a house), AR devices (automatically rendering an object in the user's space), and visually-grounded chatbots with common sense.

\looseness=-1 How can we build vision systems with SP prediction abilities?
%
Modern computer vision tasks have focused on classifying, localizing, and describing what is visible in an image (\eg classification, object detection, segmentation, and captioning).
Most visual representation learning approaches, \eg CLIP~\cite{radford2021learning,ilharco_gabriel_2021_5143773,Zhai2023SigLIP,Sun2023EVACLIPIT,Li2023AnIS}, use losses that encourage the learned representations to capture what is shown in the image but are not designed to be used to answer queries about the invisible in the image \emph{zero-shot}; the visual context generated by these models is, however, extremely valueable and we use CLIP as the visual backbone in this work. 
%
%
%
%
Recent advances in vision-and-language (VLM) foundation models has made some progress in this direction.
We can ask VLMs questions that require reasoning about the invisible, conditioned on visual context to infer the answers to a question.
However, existing VLMs are still in early stages and struggle to answer queries that require precise localization in pixel space as shown in our experiments (see~\cref{sec:experiments}).

In this paper, we study the problem of \underline{S}emantic \underline{P}lacement (\SP) of objects in images. 
In particular, given an image (\eg showing a living room) and name of an object ("cushion"), a vision system is tasked to predict a pixel-level mask highlighting semantically-meaningful regions (referred as \SP masks) in an image where that object could be placed or is likely to be placed by humans (\eg a couch). 
%
Learning to predict \SP masks is hard, since the target object is typically not visible in the given image.
Datasets for image description are typically constructed by curating relevant images (e.g. via image search with object names) and asking humans to annotate the contents of the image; neither of those two steps are straightforward for objects \emph{not} present in the image. 

To overcome this challenge, we propose to operate in the opposite direction -- specifically, we start with an image of an object in context (which is easy to find online) and \emph{remove} that object from the image via inpainting~\cite{suvorov2021resolution,rombach2021highresolution}. 
This automated pipline converts unstructured web data into a a dataset comprising pairs of images with/without the object at scale \emph{without expensive human annotation}. 
%
%
%
However, inpainting models are not perfect. We find that \SP prediction models, when trained on inpainted images, tend to latch onto inpainting artifacts. This leads to high performance on inpainted images, but lower performance on real images.
%
To remedy this, we propose a novel data augmentation method, combining results from multiple inpainting models, diffusion based augmentations, and common data augmentations (refer~\cref{sec:dataset} section for more details). 
%
Using this automated pipeline, we generate a large SP dataset using real world images from LAION~\cite{schuhmann2022laion}, including ${\sim}1.3$ million images across $9$ object categories.

\looseness=-1 We propose a simple method for \SP mask prediction by using a frozen CLIP~\cite{radford2021learning} backbone with a language conditioned UNet~\cite{ronneberger2015u} decoder inspired by LingUNet~\cite{lingunet} and CLIPort~\cite{shridhar2021cliport}, in Sec.~\ref{sec:model}.
First, we pretrain the  CLIP-UNet model on images from our SP dataset and then finetune on a small high-quality image dataset of ${\sim}80k$ synthetic images collected from synthetic HSSD~\cite{khanna2023hssd} scenes, where inpainting is unnecessary as objects can be removed programmatically from the underlying 3D scenes.
%
We find finetuning on this small but high-quality dataset with ground truth object placement annotations improves performance of our CLIP-UNet baseline and enables better generalization to both real and synthetic images.

For evaluation we use $400$ real world images from LAION~\cite{schuhmann2022laion} and ${\sim}18k$ from HSSD~\cite{khanna2023hssd} scenes.
We find that CLIP-UNet outperforms strong baselines leveraging VLMs, including LLaVa-1.5~\cite{liu2023improvedllava} and GPT4V~\cite{OpenAI2023GPT4TR}, and methods using open-vocabulary object detection and segmentation models with placement priors coming from LLMs.
In user studies, we find that the \SP mask predicted by our method are favored $43.7\%$ times against the baselines on real images and by $31.3\%$ times on images from HSSD scenes.
%

SP mask predictions hold potential for a variety of downstream applications, including assistive agents, real-time AR rendering, and visually-grounded chatbots. In this paper, we demonstrate that SP masks predicted by CLIP-UNet enable embodied agents to perform Embodied Semantic Placement (\ESP) task in a photorealistic, physics-enabled simulated environment, Habitat~\cite{savva_iccv19,andrew_neurips21,puig2023habitat3} using Hello Robot's Stretch robot~\cite{Kemp2022StretchRobot}. 
In \ESP, an agent is spawned at a random location in an indoor environment and is tasked with placing an instance of a target object
category at a semantically meaningful location with access to robot observations (RGB, Depth, and pose) and \SP masks from a \SP model.
Using \SP masks predicted by our CLIP-UNet model, agent achieves a $12.5\%$ success rate on $8$ categories when evaluated in $10$ unique indoor scenes over $106$ episodes.
While the absolute success is indeed low, we note that majority of failures ${\sim}80\%$ for downstream \ESP task are due to imperfect control policy for object placement and fine-grained navigation, which is orthogonal to the focus of this work.
We show a qualitative example of a placement prediction by our agent for object category `cushion' while performing the task in Fig.~\ref{fig:task} (c).
%
%

In summary, our contributions include: (1) a novel task called Semantic Placement (\SP), (2) an automated data curation pipeline leveraging inpainting and object detection models to supervise an end-to-end \SP prediction model using real-world data, (3) a novel data augmentation method to alleviate overfitting to inpainting artifacts, and (4) our approach generates \SP predictions which generalize well to the real-world and enable downstream robot execution.

\begin{figure*}[h]
  \centering
    \includegraphics[width=1\textwidth]{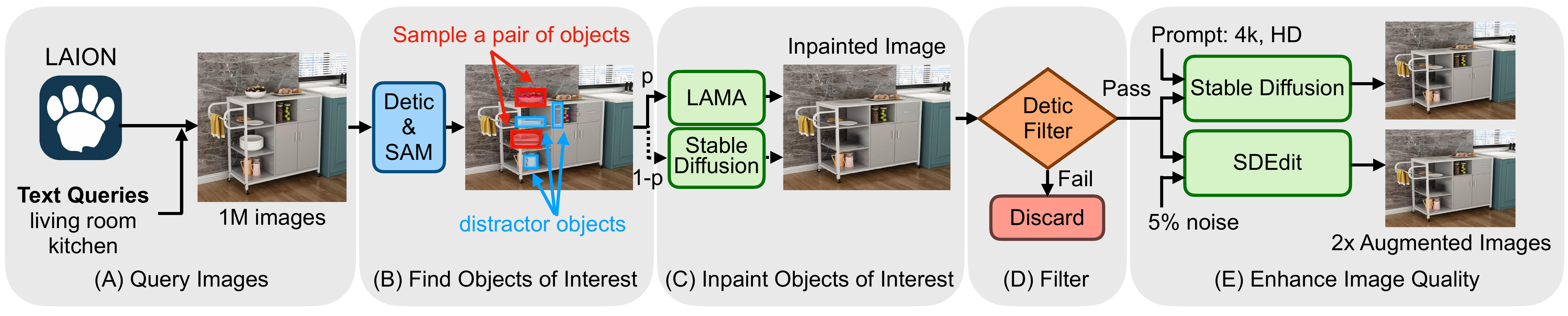}
    \caption{\textbf{Automatic Training Dataset Generation Pipeline Utilizing Foundation Models and Web Data.} 
    Our pipeline consists of five steps.
    (A) \textit{Query Images}: we collect raw images from LAION~\cite{schuhmann2022laion} using sample text queries such as `living room' shown in the leftmost panel. (B) \textit{Find Objects of Interest}: we employ Detic~\cite{zhou2022detecting} and SAM~\cite{kirillov2023segany} to identify the segmentation masks of objects of interest. (C)\textit{Inpaint Objects of Interest}: we use inpainting models to remove the objects of interest from the images. (D) \textit{Filter}: we discard images where impainting failed by attempting to detect inpainted objects. (E) \textit{Enhance Image Quality}: we leverage Stable Diffusion img2img~\cite{rombach2021highresolution} and SDEdit~\cite{meng2022sdedit} to enhance the quality of the generated images, which is crucial for training our Semantic Placement model.}
    \vspace{-1em}
    \label{fig:data_gen}
\end{figure*}

\section{Related Work}
\label{sec:related_work}

\textbf{Object Affordance Prediction from Common-sense Reasoning.} 
Object affordance~\cite{luo2022learning,chao2015mining,hadjivelichkov2023one}  is defined as a function that can map images of object to potential interactions that are possible, like holdable, pushable, liftable, placeable, etc. Learning such a function requires learning characteristics of a object based on visual appearance, semantics, or physical characteristics. 
In contrast, we are interested in Semantic Placement task which requires reasoning about placement of an object that is \textit{not present} in the image using the context and semantics of what is present in the image.
Prior works~\cite{kant2022housekeep,sarch2022tidee,wu2023tidybot} have leveraged LLMs to extract object affordances in the form of states like whether a object is misplaced or is it a receptacle \ie where you can place another objects, to build agents to tidy up a indoor environment.
LLMs~\cite{brown2020language,devlin2018bert,liu2019roberta,thoppilan2022lamda,raffel2020exploring,wei2021finetuned} and VLMs~\cite{zellers2022merlot,zellers2021merlot,zhang2023gpt4roi,su2019vl,alayrac2022flamingo,li2023blip,li2022blip,zhu2023minigpt} demonstrate strong common-sense reasoning about object affordances based on visual appearance or semantics, however they seldom output \SP mask/heatmap predictions with sufficient granularity to allow for the precise placement localization required for downstream tasks.

\textbf{Learning Visual Affordances for Object Placement.} 
Also related to \SP is prior work on object affordances~\cite{luo2022learning,chao2015mining,hadjivelichkov2023one} for tasks such as tabletop manipulation~\cite{chu2019toward,lin2023mira,shridhar2021cliport,zeng2021transporter}, articulated manipulation~\cite{wang2022adaafford,mo2021where2act,ning2023where2explore,geng2022end, zhao2022dualafford}, dexterous grasping~\cite{li2023gendexgrasp}, and interactions between embodied agents and environments~\cite{nagarajan2020learning}. 
These works focus on learning affordances for manipulation about where to interact and how to interact with the object by leveraging labelled simulation data, exocentric images, and limited real world robot data.
In contrast, our work focuses on predicting \textit{plausible} locations for placing objects which are \textit{not present} in an image based on visual context by leveraging automatically generated large scale labelled data.
The problem we explore is more closely aligned with the concept of learning object-object affordances~\cite{sun2014object,zhu2015understanding,mo2022o2o}, which includes the challenge of placing objects within/on the receptacles. Perhaps the most similar to our work is O2O~\cite{mo2022o2o} which predicts 3D affordances maps using point cloud inputs. The O2O model was trained with data collected through simulated interactions, resulting in more geometry-aware affordance predictions, with limited generalization ability. In comparison, we propose learning a \SP model using both images in the wild~\cite{schuhmann2022laion} and a high-quality simulation environment~\cite{khanna2023hssd} which leads to better generalization ability.
Similar to our method, recent approaches also propose learning visual affordances from natural images~\cite{bharadhwaj2023visual}, human-captured videos~\cite{bahl2023affordances}, or images paired with synthesized interactions~\cite{ye2023affordance,yu2023scaling}. However, these works focus on learning affordances for what is \textit{present} in the image, in contrast, we study learning placement localization for objects that are \textit{not present}.


\section{Approach}
\label{sec:approach}

We introduce our dataset generation pipeline in Sec.~\ref{sec:dataset}, HSSD finetuning dataset in~\cref{sec:synthetic_dataset} and describe our \SP model and learning procedure in Sec.~\ref{sec:model}.

\subsection{Dataset Generation}
\label{sec:dataset}


To collect paired data for training (referred as LAION-SP) the \SP model, we propose leveraging recent advances in open-vocabulary object detectors, segmentation models, and image inpainting models. With these powerful off-the-shelf ``foundation'' models, we can generate paired training data at scale using images in the wild. Fig.~\ref{fig:data_gen} shows our automated data generation pipeline,
including five steps: \textit{Query Image}, \textit{Find Objects of Interest}, \textit{Inpaint Objects of Interest}, \textit{Filter}, and \textit{Enhance Image Quality}. At the end of the pipeline, each output image is paired with object categories and includes masks showing where such categories can be placed. Details follow below.

\noindent \textbf{(A) \textit{Query Image.}} First, we gather 1M indoor images from the LAION dataset by using text queries such as `living room', `bedroom', and `kitchen' to filter out irrelevant images \ie images not from houses. 
%

\noindent \textbf{(B) \textit{Find Objects of Interest.}} Next, for each image we use Detic~\cite{zhou2022detecting}, an open vocabulary object detector, to detect objects of interest for our task. We use $9$ target object categories in this paper, specifically  \texttt{Plotted Plant}, \texttt{Lamp}, \texttt{Cushion}, \texttt{Vase}, \texttt{Trash Can}, \texttt{Toaster}, \texttt{Table Lamp}, \texttt{Alarm Clock}, and \texttt{Laptop}. For each detected instance, we generate a segmentation mask using SAM~\cite{kirillov2023segany}.
We use SAM masks instead of Detic masks as they are fine-grained and result in better inpainting performance. For information on how we prompt SAM and Detic to generate segmentation masks, see~\cref{sec:data_supp}.
%

\noindent \textbf{(C) \textit{Inpaint Objects of Interest.}} Using the detection results, we pass the segmentation masks of instances of a sampled object category and original image to one of the two inpainting models (each sampled with 50\% probability), LAMA~\cite{suvorov2021resolution} or Stable Diffusion~\cite{Rombach_2022_CVPR}, to generate an inpainted sample.
Specifically, we randomly sample a few instances of a target object category and $1$-$4$ distractor objects of different category for inpainting.
We add distractor instances to make the task of \SP prediction more challenging as the model cannot simply predict the, possibly only, free space. This also helps prevent the model from  overfitting to inpainting artifacts.
%
%

\noindent \textbf{(D) \textit{Filter.}} Inpainting models are imperfect and we need strict validation mechanisms to check if inpainting was successful or not. 
To do so, we use 2D instance matching between original and inpainted images using the detections from Detic~\cite{zhou2022detecting}. 
%
Specifically, if we find an object instance post-inpainting with IOU greater than $90\%$ with an instance from original image, the inpainting model failed and we discard the generated result. 
%
%
All samples that pass the validation check are kept as part of training dataset. 
%

\noindent \textbf{(E) \textit{Enhance Image Quality.}} In our initial experiments, we found that training the SP model directly using the dataset generated by the \textit{Filter} step leads to overfitting.\footnote{The model trained on the inpainted images without quality enhancement (\ie, Step E) yields ${\sim}0$ TP zero-shot evaluating on HSSD dataset.}
The model quickly latches onto the artifacts introduced from the inpainting models. To mitigate this issue, we generate two augmented versions of each inpainted image with the help of diffusion models.
To create the first augmented variant, we add $5\%$ Gaussian noise to the image and use SDEdit~\cite{meng2022sdedit} to denoise the image similar to Affordance Diffusion~\cite{ye2023affordance}.
To create the second variant, we feed the inpainted image to Stable Diffusion img2img~\cite{rombach2021highresolution} model and prompt it with `high resolution, 4k' which, in practice, results in small object texture changes.
%
We find this acts as regularization and helps avoid overfitting on inpainting artifacts during training.

\begin{figure}[t]
  \centering
    \includegraphics[width=0.4\textwidth]{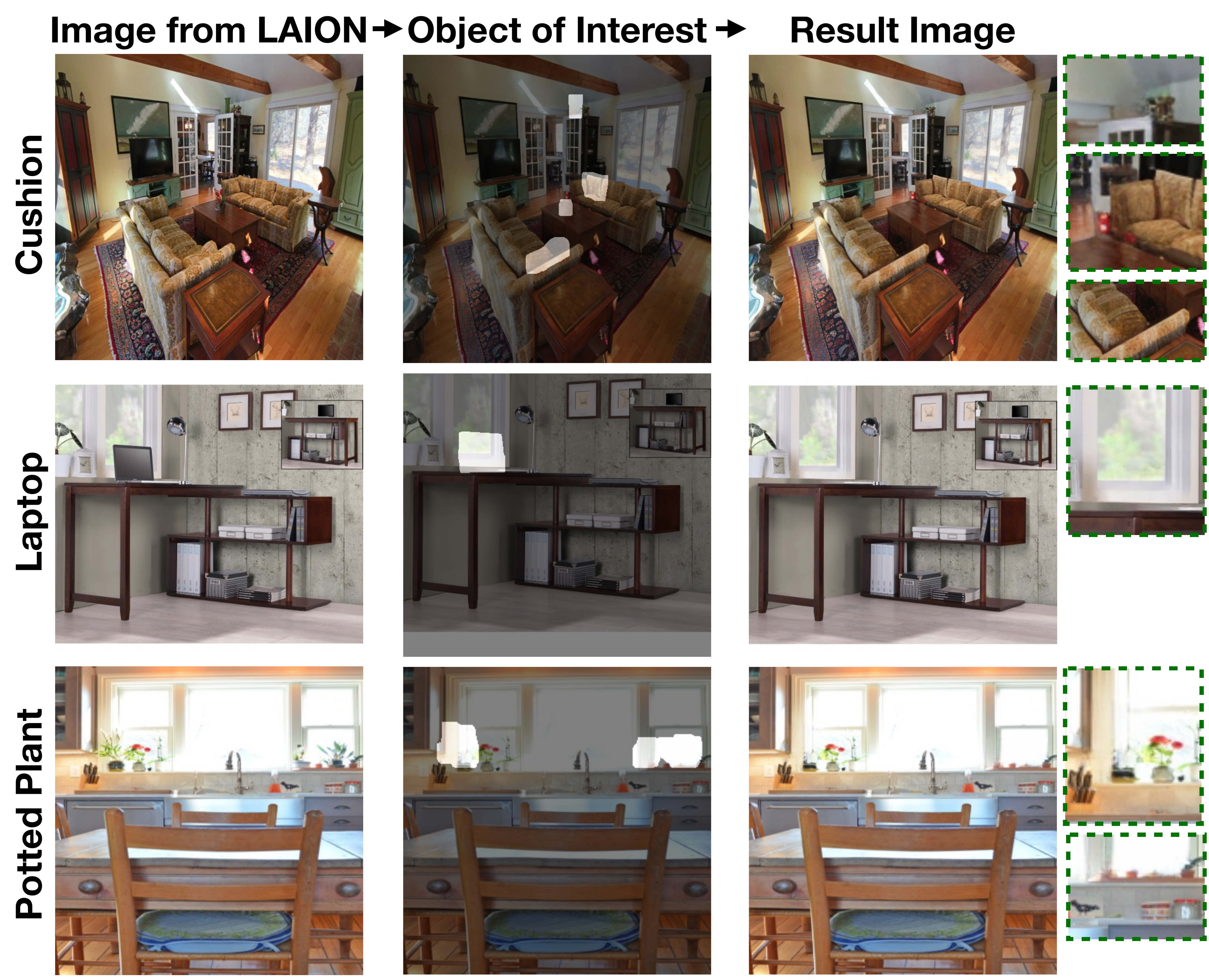}
    \caption{\textbf{Qualitative Examples of Generated Images}. We present three examples of \texttt{Cushion}, \texttt{Laptop}, and \texttt{Potted Plants}, which include raw images queried from LAION (left), identified objects of interest and their segmentation masks obtained from SAM (middle), and the result images after \textit{Inpainting}, \textit{Flitering}, and \textit{Quality Enhancement} steps (right). For clarity, we have magnified the inpainted regions, highlighted in green dotted boxes.
    }
    \vspace{-1em}
    \label{fig:data_gen_results}
\end{figure}

For each image processed in this way, we are left with two augmented and inpainted images, both paired with \SP annotations for an object category corresponding to the SAM masks generated at the beginning of the processing to form training samples. In total, we generate $1{,}329{,}186$ images with an object category and its corresponding SAM mask from $48{,}728$ unique images queried from the LAION dataset. In Tab.~\ref{tab:images_per_category}, we show the number of generated images per object category. 
Fig.~\ref{fig:data_gen_results} showcases three qualitative examples generated by our dataset generation pipeline, including \texttt{Cushion}, \texttt{Laptop}, and \texttt{Plotted Plant}.
In addition to LAION-SP training dataset of ${\sim}1.3M$ images we also create a dataset of $400$ unseen original images from LAION for our evaluation referred as LAION-SP Val dataset.


\begin{table}[h]
\centering
\resizebox{1\columnwidth}{!}{
\begin{tabular}{|l|r|r|r|r|r|}
\hline
\textbf{Category} & \textbf{Potted Plant} & \textbf{Lamp} & \textbf{Cushion} & \textbf{Vase} & \textbf{Trash Can} \\ \hline
\# Images & 207,366 & 320,922 & 323,541 & 417,591 & 13,353 \\ \hline
\end{tabular}
}
~\\[0.1em]
\resizebox{0.9\columnwidth}{!}{
\begin{tabular}{|l|r|r|r|r|r|}
\hline
\textbf{Category} & \textbf{Toaster} & \textbf{Table Lamp} & \textbf{Alarm Clock} & \textbf{Laptop} \\ \hline
\# Images & 23,928 & 5,559 & 14,496 & 2,430 \\ \hline
\end{tabular}
}
~\\[0.1em]
\caption{Number of Images per Category in LAION-SP Dataset.}
\label{tab:images_per_category}
\vspace{-0.5em}
\end{table}

\subsection{Synthentic Images}
\label{sec:synthetic_dataset}

\begin{figure*}[t]
  \centering
  \vspace{-1em}
    \includegraphics[width=0.9\textwidth]{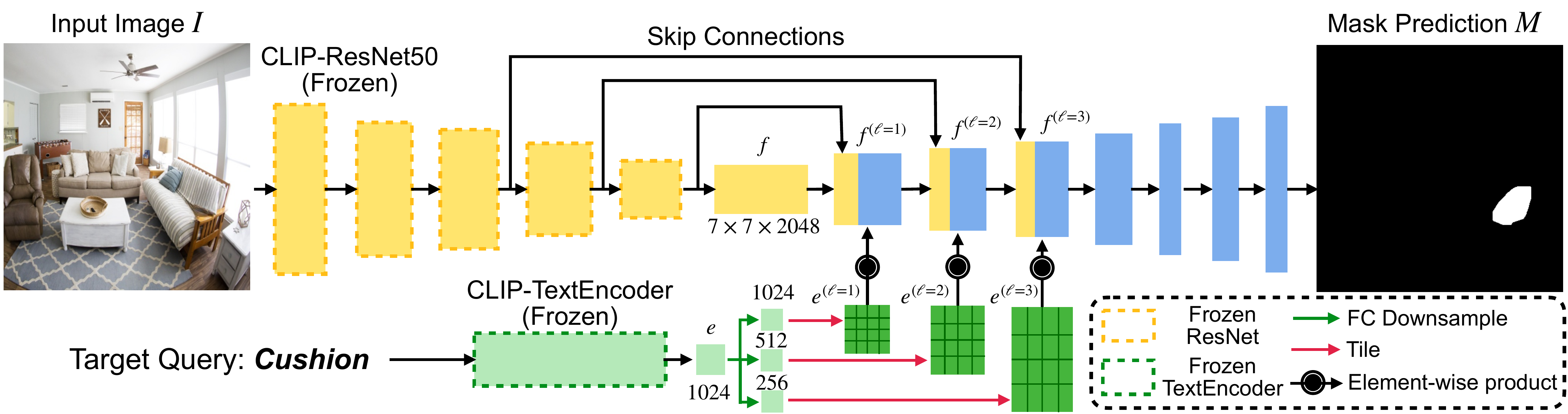}
    \caption{\textbf{CLIP-UNet for the SP task}. Inspired by CLIPort~\cite{shridhar2021cliport}, we first encode the input image $I$ into a feature sensor $f$, and encode the target object category $q$ into an embedding $e$. Further downsampling and tiling ensure that the target embedding matches the dimension of the feature tensors $f^{(\ell)}$ at the first three decoder layers. We then use an element-wise product to combine the target embedding $e^{(\ell)}$ and the feature tensor $f^{(\ell)}$ to achieve semantic conditioning. Similar to LingUNet~\cite{lingunet}, we add skip-connections for these three layers. Finally, CLIPort outputs a mask prediction on the image, indicating the optimal region to place the given target object.}
    \vspace{-1em}
    \label{fig:architecture}
\end{figure*}

For finetuning, we collect a small high-quality image dataset from synthetic HSSD~\cite{khanna2023hssd} scenes, a synthetic indoor environment dataset comprising $211$ high-quality 3D scenes, containing $18{,}656$ models of real-world objects.
To generate the dataset using HSSD scenes inpainting is unnecessary as the objects can be removed programatically from underlying 3D scenes and the image can be re-rendered from the same viewpoint using Habitat~\cite{savva_iccv19,andrew_neurips21} simulator.
%
%
%
Using $135$ training scenes we generate ${\sim}80k$ training images across $8$ object categories.
Similarly, using $33$ unseen evaluation scenes we generate a dataset of ${\sim}18k$ images for evaluation with the same $8$ object categories.
Additional details on image generation and viewpoint sampling is in App.~\ref{sec:hssd_dataset}.

\subsection{Learning Object Placement Affordance}
\label{sec:model}

To learn an \SP mask prediction model, we use the dataset generated from~\cref{sec:dataset}.
The inputs to the \SP model include an RGB image $I$ in $H{\times} W {\times} 3$ size and a target object category $q$ in text. The model outputs an affordance mask $M$, size $H{\times} W{\times} 1$, conditioned on the target object.
%
\cref{fig:architecture} shows the architecture of our proposed CLIP-UNet model.
Inspired by CLIPort~\cite{shridhar2021cliport}, we use a frozen ResNet50~\cite{he_cvpr16}, pre-trained by CLIP~\cite{radford2021learning}, to encode the input image $I$ into a feature tensor $f$ up until the penulitmate layer $\mathbb{R}^{7\times7\times2048}$.
The decoder then upsamples the feature tensor $f$ to $f^{(\ell)} \in \mathbb{R}^{H_\ell \times W_\ell \times C_\ell}$ at each layer $\ell$ and, at the end, produces a mask $M \in \mathbb{R}^{H \times W \times 1}$, where $0 \leq M[i, j] \leq 1$.

To encode the target object category $q$, we use CLIP pretrained transformer based sentence encoder to construct a target embedding $e \in \mathbb{R}^{1024}$.
%
To condition the decoding process with the target embedding, we first downsample it to $\bar{e} \in \mathbb{R}^{C_{\ell}}$ and then tile it to match the dimension of feature tensor $f^{(\ell)}$ at layer $\ell$ in the decoder:
$\bar{e} \rightarrow \bar{e}^{(\ell)} \in \mathbb{R}^{H_\ell \times W_\ell \times C_\ell}$, where $C_{\ell}=\{1024, 512, 256\}$ and $\ell\in\{1, 2, 3\}$.
Then, we use the tiled target embedding to condition the visual decoder layers through an element-wise product. As CLIP utilizes contrastive loss on the dot-product aligned features from pooled image features and language embeddings, the element-wise product allows us to leverage this learned alignment while the tile operation preserves the original dimensions of visual features.
Inspired by LingUNet~\cite{lingunet}, we apply this language conditioned operation to the first three upsampling layers right after the feature tensor $f$ produced by the frozen ResNet. 
%
Moreover, following UNet~\cite{ronneberger2015u}, we add skip connections to decoder layers from the corresponding layers in ResNet encoder. In this way, the model preserves different levels of semantic information from input image.

\textbf{Training Details}. We train our CLIP-UNet model in two stages. First, we pretrain our model using the LAION-SP dataset generated in Sec.~\ref{sec:dataset}, containing $1.3$M images across $9$ categories for $10$ epochs using \textit{dice loss}~\cite{sudre2017generalised}.
During pretraining, in addition to diffusion model augmented images, we also use common data augmentations, such as gaussian blurring, additive gaussian noise, horizontal flipping, and color jitter to mitigate inpainting artifacts. 
Next, we finetune the LAION-SP pretrained model using a small, high-quality, dataset generated using synthetic HSSD scenes~\cite{khanna2023hssd,savva_iccv19,andrew_neurips21} mentioned in~\cref{sec:synthetic_dataset}.
As the HSSD image dataset is generated using a simulator we can manipulate the scene to render images with/without object images without introducing any artifacts that models can latch on to.
This two-stage training improves performance of our CLIP-UNet model and enables better generalization to both real and synthetic images as shown in~\cref{sec:experiments}.

\section{Evaluation Metrics}
\label{sec:metrics}

\begin{figure}[t]
  \centering
    \includegraphics[width=0.8\linewidth]{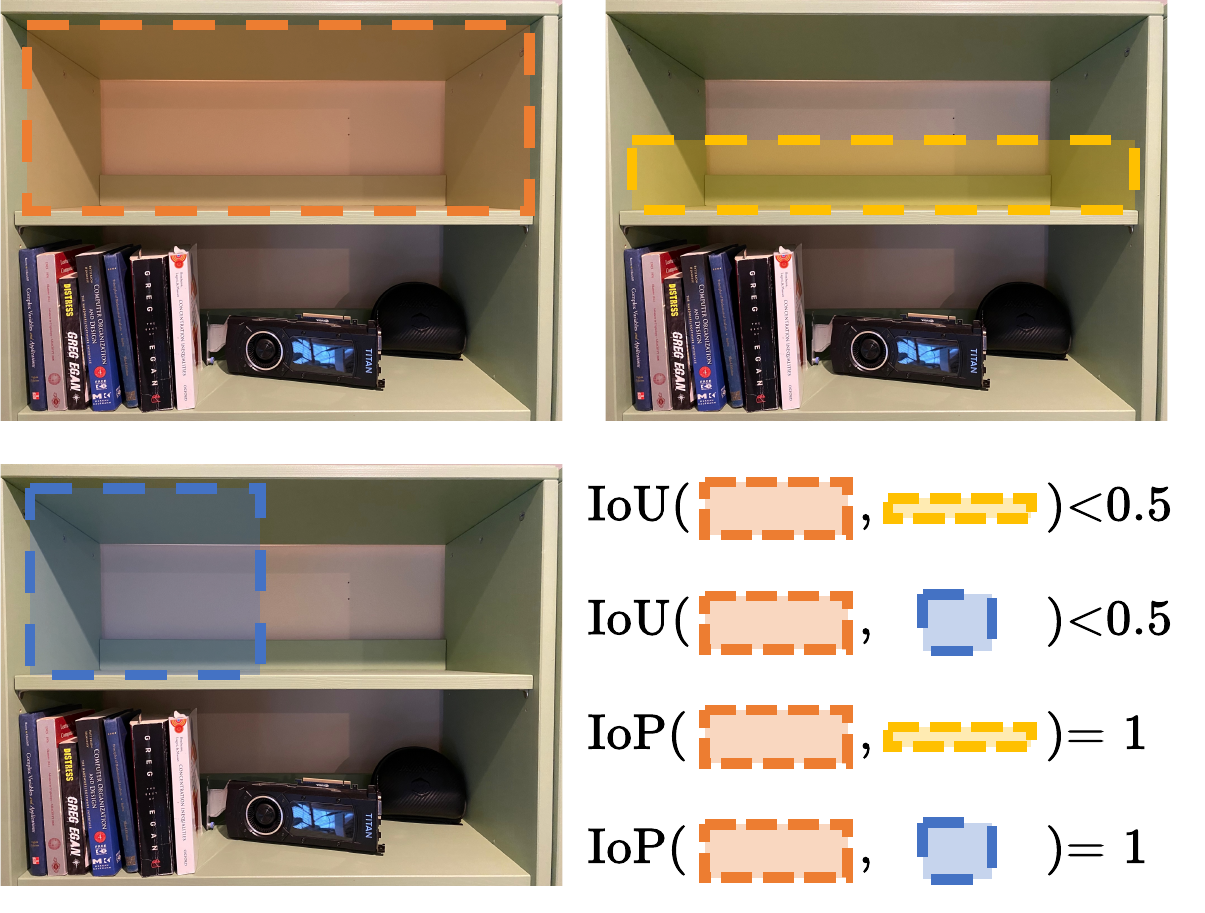}
    \caption{\textbf{\textsc{IoU} \emph{v.s.} \textsc{IoP}.} Top left: a hypothetical ground-truth (GT) \SP region for objects of type ``book''.
    Top right \& bottom left: two possible \SP predictions. Both predicted regions are high-quality and should be considered true-positives. The IoU for these
    predictions is, however, $<0.5$ as the \textsc{IoU} normalizes by the large GT region. The \textsc{IoP}, however, only normalizes by the predicted mask's size and thus
    is equal to 1 for both predicted regions.}
    \vspace{-2em}
    \label{fig:iou-vs-iop}
\end{figure}

In this section, we propose metrics for evaluating \SP prediction performance. Before defining these metrics, we will begin by defining what we mean by true/false positive (TP/FP) and true/false negative (TN/FN) \SP predictions.

\textbf{Preliminaries.} Consider an image $I$, an object type query $q$, and an (exhaustive) set of $\{0, 1\}$-valued ground-truth disjoint regions $r_1, ..., r_K\in\{0,1\}^{H{\times}W}$ describing the locations where objects of type $q$ can be placed in the image $I$. Let the model produced region predictions be denoted by $\hat{r}_1,...,\hat{r}_L\in \{0,1\}^{H{\times}W}$. Intuitively, we would like a predicted region  $\hat{r}_j$ to be considered a true positive (TP), if it ``overlaps sufficiently'' with some GT region $r_i$. Measuring region overlap is commonly achieved, \eg in the semantic segmentation and object detection literature~\cite{Detectron2018,lin2014microsoft,kirillov2023segment}, using the intersection-over-union (\textsc{IoU}) metric, $\textsc{IoU}(r, r') = r \bcdot r' / (r\bcdot r + r'\bcdot r'-r\bcdot r')$ where $\bcdot$ denotes the usual dot product. The \textsc{IoU} works well when one wishes to enforce that two regions overlap \emph{exactly}; for \SP prediction, however, requiring exact overlap is too restrictive as it normalizes by too large of a region, see Fig.~\ref{fig:iou-vs-iop}. Instead we use the intersection-over-prediction $\textsc{IoP}(r,r') = r\bcdot r' / (r'\bcdot r')$ which normalizes only by the size of the predicted region $r'$.

That is, we say that  $\hat{r}_j$ is a TP if there exists some $r_i$ such that $\textsc{IoP}(\hat{r}_j, r_i) \geq T$ where $T\in[0,1]$ is some threshold value (for us, $T=0.5$). We say that $\hat{r}_j$ is a FP if there is no $r_i$ with $\textsc{IoP}(\hat{r}_j, r_i) \geq T$.
Importantly: TPs are counted with respect to the ground truth region $r_i$ while FPs are counted with respect to the predicted region $\hat{r}_j$.
This means that that if the model predicts multiple regions $\hat{r}_j$ which all correspond to a single $r_i$, then these multiple regions will be counted as only a single TP.
Additionally, number of FN equal to number of GT regions $r_i$ not covered by any predicted region $\hat{r}_j$. 

\textbf{Precision and recall.} Given the above, we can now define the usual recall and precision metrics for an image $I$ as $\text{Precision}(I) = \frac{\# TP}{\# TP + \# FP}$ and $\text{Recall}(I) = \frac{\#TP}{\#TP + \# FN} $. When reporting metrics on our evaluation sets, we report the average precision and recall over all images. If an image $I$ has no GT masks, then $\text{Recall}(I)$ is not well-defined and so we do not include such images when computing the average. 
We compute these metrics only on HSSD dataset as these require access to accurate GT region annotations.

%

\textbf{Receptacle priors.} One important facet of \SP prediction is an understanding of the relationship between receptacle types and the objects that are typically placed upon them. For instance, you will almost always find a plunger on the floor and not on a dining table. Indeed, it is exactly these types of receptacle relationships that some previous work, \eg \cite{sarch2022tidee,kant2022housekeep,wu2023tidybot}, have focused upon.
In order to measure the model's ability to encode such priors, we introduce the receptacle surface precision (RSP) and receptacle surface recall (RSR) metrics. To compute these metrics, we first, for each object type query $q$, curate a collection of receptacle types that such an object is commonly found upon (see Sec~\cref{sec:metrics_supp} for more details). We then, for each image $I$ and object type query $q$, assume we have access to segmentation masks $s_1,...,s_K$ of receptacles upon which $q$ is commonly found.
Moreover, as large parts of each receptacle mask will correspond to unplaceable areas (\eg the legs of a couch) we further assume that each $s_i$ corresponds only to the areas of the receptacle that are ``placeable'', \ie have a surface normal that is pointing (approximately) upward.
In practice, computing the receptacle masks can often be done automatically by leveraging simulated environments in which object
categories and geometry are known (e.g. HSSD), or by using open-vocabulary object detectors and depth maps for real world images.
As the results from open-vocabulary detectors and depth maps are noisy we only report these metrics on HSSD image dataset where we have access to ground truth.
We can then compute the RSR and RSP just as above by replacing the GT regions $r_i$ with the surface grounded receptacle segmentation masks $s_i$.
%

\textbf{Target Precision (TrP)}. To quantify precision of \SP models at localizing possible ground truth placements we compute the Target Precision (TrP) metric.
%
%
To compute TrP, we programatically compute the GT placement masks for an object category from HSSD scenes and use these as GT regions for computing the precision metrics.

\textbf{Human preference (HP).} To understand how humans judge our baselines outputs, we require human annotators to rank each model's \SP predictions from most preferred to least preferred when shown predictions from $5$ models described in~\cref{sec:experiments}.
We then report the \% of time that these annotators rank each model's predictions as the best, \ie ranked above all others, among $5$ \SP predictions. Further details in App.~\ref{sec:metrics_supp}.

\vspace{-8pt}
\section{Experiments}
\label{sec:experiments}
\vspace{-2pt}

\subsection{Semantic Placement Evaluation}
\label{sec:sp_baselines}

\begin{figure*}[h]
  \centering
    \includegraphics[width=0.9\textwidth]{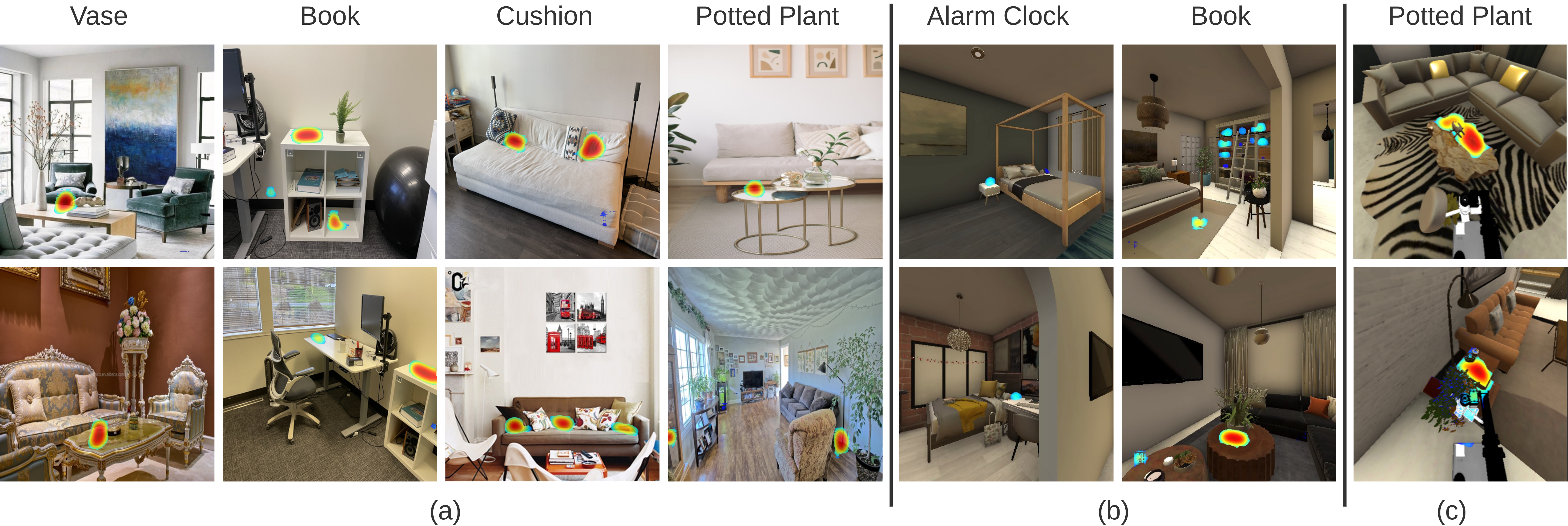}
    \caption{Qualitative examples of \SP masks predicted by our CLIP-UNet model pretrained on LAION-SP dataset and finetuned on HSSD images. (a) shows evaluation results on real image dataset from LAION~\cite{schuhmann2022laion},
    (b) shows results on images from HSSD dataset~\cite{khanna2023hssd}, and (c) shows results of placement predicted while evaluating tidying robot on Embodied Semantic Placement (\ESP) task.
    }
    \vspace{-2.0em}
    \label{fig:qualitative_example}
\end{figure*}

Here we present evaluation results on two image datasets: 1.) LAION-SP Val: $400$ real images collected from LAION~\cite{schuhmann2022laion},
2.) HSSD Val: $18k$ images from unseen HSSD scenes~\cite{khanna2023hssd}.
First, we describe baselines used for evaluation:

\textbf{LLM + Detector}. In this baseline we leverage common-sense priors from LLMs to find target receptacles for a particular object and use a open-vocabulary detector, Detic~\cite{zhou2022detecting}, to localize the receptacle in the image.
First, for each of the $9$ object categories in the dataset we prompt an LLM for common receptacle categories on which each object is found in indoor environment.
Next, during evaluation we use Detic to localize the segmentation mask of all valid receptacles for a object category in an image.

\textbf{LLaVA}. VLMs like LLaVa~\cite{liu2023improvedllava} connect vision encoders to LLMs which exhibits general purpose vision-and-language understanding.
%
%
To evaluate LLaVA on \SP, given the input image we prompt it to output normalized bounding box coordinates to localize 
a placement area.
%
Next, we convert the predicted normalized bounding box to a binary segmentation mask to use as the \SP mask for downstream applications.
Refer~\cref{sec:baseline_supp} for the prompt and sample predictions.

\textbf{GPT4V~\cite{openai2023gpt4}}. Similar to LLaVA~\cite{liu2023improvedllava}, GPT4V is a multimodal LLM with remarkable vision-and-language understanding.
To evaluate GPT4V, we pass the input image and prompt it to output normalized bounding box coordinates to localize a placement area which is then converted to a binary segmentation mask to use as the \SP mask.
Refer~\cref{sec:baseline_supp} for the GPT4V prompt and sample predictions.

\textbf{Ours (HSSD)}. Variant of our CLIP-UNet model described in~\cref{sec:model} trained only on data collected from HSSD scenes \ie no pretraining on the LAION-SP dataset from~\cref{sec:dataset}.

\textbf{Ours (LAION-SP$\rightarrow$HSSD)}. Our CLIP-UNet model from~\cref{sec:model}: first pretrained on the LAION-SP dataset and then finetuned on a small image dataset from HSSD scenes.

\begin{table}[h]
    \centering
    \resizebox{1.0\linewidth}{!}{
        \begin{tabular}{@{}cllcrcrrcrr@{}}
            \toprule
            & & & \textsc{LAION-SP Val} & & \multicolumn{4}{c}{\textsc{HSSD Val}} \\
            \cmidrule{4-4} \cmidrule{6-9}
            & & Method & HP $(\mathbf{\uparrow})$ & & HP $(\mathbf{\uparrow})$ & TrP $(\mathbf{\uparrow})$ & RSP $(\mathbf{\uparrow})$ & RSR $(\mathbf{\uparrow})$ \\
            \midrule
            \\[-10pt]
            & \rownumber & LLM + Detector & $21.5$ &  & $29.8$ & $10.1$ & $\mathbf{41.0}$ & $38.2$ \\
            & \rownumber & LLaVA & $4.9$ & & $6.8$ & $0.0$ & $26.3$ & $\mathbf{43.4}$ \\
            & \rownumber & GPT4V & $9.4$ & & $8.3$ & $-$ & $-$ & $-$ \\
            \midrule
            & \rownumber & Ours (HSSD) & $20.1$ & & $23.0$ & $16.2$ & $26.6$ & $36.5$ \\
            & \rownumber & Ours (SP $\rightarrow$ HSSD)  & $\mathbf{43.7}$ & & $\mathbf{31.3}$ & $\mathbf{18.5}$ & $24.9$ & $35.3$ \\
            \bottomrule
            \end{tabular}
    }
    \vspace{5pt}
    \caption{\textbf{SP evaluation on LAION-SP and HSSD validation splits.} We show evaluation results of our model (rows 4-5), 
    Prior + Detector, and VLM baselines. HP denotes \underline{H}uman \underline{P}reference, TrP denotes \underline{T}a\underline{r}get \underline{P}recision, RSP denotes \underline{R}eceptacle \underline{S}urface \underline{P}recision, and RSR denotes \underline{R}eceptacle \underline{S}urface \underline{R}ecall. We use $\uparrow$ to indicate that larger values are preferred.
    }
    \label{tab:main_results}
\end{table}

\textbf{Results}. ~\cref{tab:main_results} reports results of evaluating methods on the LAION-SP and HSSD evaluation datasets.
%
In our human preference study, our method (row 5) is favored the most by a large margin on real world images, and modestly in simulated images, when asked to rank predictions from all $5$ baselines from~\cref{tab:main_results}.
This demonstrates the effectiveness of using web data for pretraining our CLIP-UNet model.
%
In addition to human preferences, we also conduct quantitative evaluation using metrics from~\cref{sec:metrics}.
Our method outperforms a strong baseline that uses an LLM prior and object detector Detic (row 1) on target precision (TrP) by $8.4\%$, is comparable in the RSR metric, and performs worse on RSP metrics.
This shows that the CLIP-UNet has higher precision at localizing high-quality target placements available in the HSSD dataset, but has poor precision, compared to the Prior${+}$Detector baseline, when localizing all possible visible receptacles in the image.
%
We hypothesize that this low precision is caused by false positive predictions in the vicinity of receptacles not grounded to appropriate surfaces.

Our method (row 5) also outperforms both VLM baselines, \ie LLaVA (row 2), significantly on TrP and achieves comparable performance on the RSP metric.
In addition, our method also outperforms GPT4V on human preference evaluation by $34.3\%$ on real images and $23.0\%$ on HSSD images.
Due to current GPT4V API limits, quantitative evaluation on $18k$ images from HSSD val split would've taken 180 days so we could not show quantitative results in row 3.
After some preliminary analysis of results of the VLMs we find that, when tasked to output the placement location as language in addition to bounding box coordinates, these VLMs do a good job at giving reasonable responses but fail to precisely localize the output in the image space.
More details in App{.}~\ref{sec:results_supp}.
These results demonstrate the difficulty of SP prediction and highlight that there's still scope for improvements in general-purpose VLMs.
Next, we compare our method (row 5) against a CLIP-UNet trained only on the HSSD dataset (row 4), and we find that LAION-SP pretraining helps significantly in improving generalization performance of CLIP-UNet baseline. 
Specifically, we see improvements of $23.6\%$ and $8.3\%$ in human preferences on real and HSSD images, respectively, and $2.3\%$ improvement in target precision.

\begin{table}[H]
    \centering
    \resizebox{0.8\linewidth}{!}{
        \begin{tabular}{@{}cllrcrr@{}}
            \toprule
            & & & \multicolumn{3}{c}{\textsc{HSSD Val}} \\
            \cmidrule{4-6}
            & & Method & TrP $(\mathbf{\uparrow})$ & RSP $(\mathbf{\uparrow})$ & RSR $(\mathbf{\uparrow})$ \\
            \midrule
            \\[-10pt]
            & \rownumber & Ours (LAION-SP) & $10.1$ & $23.7$ & $26.3$ \\
            & \rownumber & Ours (HSSD) & $16.2$ & $\mathbf{26.6}$ & $\mathbf{36.5}$ \\
            \midrule
            & \rownumber & Ours (LAION-SP $\rightarrow$ HSSD) & $\mathbf{18.5}$ & $24.9$ & $35.3$ \\
            \bottomrule
            \end{tabular}
    }
    \vspace{5pt}
    \caption{\textbf{LAION-SP pretraining ablations.} We show the evaluation results by training our CLIP-UNet on different datasets.
    }
    \vspace{-1em}
    \label{tab:ablate_pretraining}
\end{table}

\textbf{Effectiveness of Pretraining on the SP Dataset}.
\cref{tab:ablate_pretraining} shows results varying the CLIP-UNet training dataset.
First, evaluating the model trained on the LAION-SP dataset zero-shot on HSSD (row 1) results in $10.1\%$ TrP, $23.7\%$ RSP, and $26.3\%$ RSR.
This suggests the LAION-SP pretrained model is, in general, good at identifying correct receptacle surfaces in HSSD but does not,  as shown by low TrP numbers, perform very well in precisely localizing one of the ground truth object placements.
In contrast to training on the LAION-SP dataset, if we just train from scratch on HSSD images (row 2 vs 1) we achieve a $+6.1$ absolute improvement on TrP, $+2.9\%$ on RSP, and $+10.2\%$ on RSR. 
%
However, with small amounts of finetuning of the LAION-SP pretrained model on HSSD dataset (row 3 vs 2), we obtain our best performing model which obtains a further absolute improvement of $+2.3\%$ on TrP with comparable performance on RSP and RSR.
In addition, as shown in human preference numbers in~\cref{tab:main_results} (row 4 vs 5), pretraining on LAION-SP and finetuning on HSSD leads to overall better generalization to both sim and real images.
These results effectively demonstrate that pretraining on the LAION-SP dataset enables better generalization. See Fig.~\ref{fig:qualitative_example} for qualitative examples of our HSSD-finetuned model's predictions.

\textbf{Open-Vocab Object Detector Ablation.}
\cref{tab:detector_ablations} presents results for when varying the open vocabulary object detectors used in our LLM${+}$Detector baseline.
We compare performance on the HSSD validation split using TrP, RSP, and RSR metrics and consider three open vocabulary detectors: Detic~\cite{zhou2022detecting}, OwlViT~\cite{minderer2022simple}, and GroundedSAM~\cite{kirillov2023segany,liu2023grounding}.
Overall, we find Detic achieves the highest RSP, RSR, and comparable or better TrP compared to OwlVit and GroundedSAM.

\begin{table}[H]
    \centering
    \vspace{-0.5em}
    \resizebox{0.8\linewidth}{!}{
        \begin{tabular}{@{}cllrcrr@{}}
            \toprule
            & & & \multicolumn{3}{c}{\textsc{HSSD Val}} \\
            \cmidrule{4-6}
            & & Method & TrP $(\mathbf{\uparrow})$ & RSP $(\mathbf{\uparrow})$ & RSR $(\mathbf{\uparrow})$ \\
            \midrule
            \\[-10pt]
            & \rownumber & LLM + Detic & $10.1$ & $\mathbf{41.0}$ & $\mathbf{38.2}$ \\
            & \rownumber & LLM + OwlVit & $\mathbf{11.4}$ & $26.2$ & $26.2$ \\
            & \rownumber & LLM + GroundedSAM & $8.9$ & $35.1$ & $32.1$ \\
            \bottomrule
            \end{tabular}
    }
    \vspace{5pt}
    \caption{\textbf{Ablations of object detectors for prior based baselines.}} 
    \label{tab:detector_ablations}
    \vspace{-1em}
\end{table}

\subsection{Embodied Evaluation}
\label{sec:embodied_eval}

\begin{table}[t]
    \centering
    \resizebox{0.85\columnwidth}{!}{
        \begin{tabular}{@{}cllrr@{}}
            \toprule
            & & Baseline & Success $(\mathbf{\uparrow})$  \\
            \midrule
            & \rownumber & LLM + Detector                            & $10.5\%$ \\
            & \rownumber & LLaVA                                     & $9.0\%$ \\
            \midrule
            & \rownumber & Ours (LAION-SP $\rightarrow$ HSSD)        & $\mathbf{12.5\%}$ \\
            \bottomrule
            \end{tabular}
    }
    \vspace{5pt}
    \caption{\textbf{Embodied Semantic Place (eSP) evaluation performance on HSSD \textsc{val} split}. We evaluate each SP model from ~\cref{sec:sp_baselines} using a modular eSP policy with same hyperparameters.}
    \label{tab:embodied_main}
    \vspace{-20pt}
\end{table}

In this section, we present the results of using our CLIP-UNet (LAION-SP $\rightarrow$ HSSD) model for the downstream application of building a tidying robot.
Specifically, in this task, an agent is spawned at a random location in an indoor environment and is tasked with placing an instance of a target object category at a semantically meaningful location. We call this task Embodied Semantic Placement (\ESP).
For our experiments, we use Hello Robot's Stretch robot~\cite{Kemp2022StretchRobot} with the full action space as defined in~\cite{yenamandra2023homerobot}. 
Specifically, the observation space, shown in the Fig.~\ref{fig:embodied-eval}, includes RGB+Depth images from the robot's head camera, camera pose, arm joint and gripper states, and robot's pose relative to the starting pose of an episode.
The robot's action space comprises discrete navigation actions: \moveforward ($0.25m$),
\turnleft ($30^{\circ}$), \turnright ($30^{\circ}$), \lookup ($30^{\circ}$),
and \lookdown ($30^{\circ}$).
For manipulation, we use a continuous action space for fine-grained control of the gripper, arm extension and arm lift.
%

\textbf{Evaluation Dataset}.
For \ESP evaluation, we create a dataset consiting of $106$ episodes using HSSD scenes~\cite{khanna2023hssd}, each specified by an agent's starting pose and a target object category. These episodes span $8$ object categories across $10$ indoor environments.
An episode is successful if the agent successfully places the object on one of the semantically valid receptacle (\eg cushion on a bed or couch).
%

\textbf{Embodied Semantic Placement Policy}.
To perform the task with only robot observations and \SP mask predictions from the CLIP-UNet at each frame, we use a two-stage modular policy consisting of ``navigation'' and ``place'' policies.
The navigation policy employs frontier exploration~\cite{ramakrishnan_arxiv20}, building a top-down semantic map using Active-Neural SLAM~\cite{chaplot_neurips20}. 
At each timestep, using the camera pose and depth we project the predicted \SP masks onto a top-down placement affordance map and explore the environment for $150$ steps.
Following exploration, we utilize the placement affordance map to navigate within $0.2m$ of a placement area.
We then rerun the CLIP-UNet while the agent performs a panoramic turn, allowing for the identification of a precise placement prediction in the 2D image space. This prediction is then projected to 3D to sample a placement location.
Once a placement location is identified, an inverse-kinematics-based planner is used to place the object at the predicted location.
The policy is illustrated in~\cref{fig:embodied-eval}, refer App.~\ref{sec:embodied_eval_setup} for more details.

\textbf{Results}.
\cref{tab:embodied_main} presents the results of evaluating the eSP policy using SP mask predictions LLM${+}$Detector, LLaVa and our CLIP-UNet (LAION-SP$\rightarrow$HSSD) model on HSSD val split.
We do not evaluate GPT4V on eSP task due to API limitations, eSP policy evaluation requires running inference using GPT4V after each robot action which amounts to a total of ${\sim}53k$ frames for full evaluation.
We find our CLIP-UNet eSP policy achieves a $12.5\%$ success on the \ESP task across $10$ indoor environments, outperforming LLaVa and LLM+Detector eSP baselines by $2-3.5\%$ on task success.
We observe that our CLIP-UNet \ESP agent can effectively reason about appropriate object placements in these settings. For example, in a living room scenario near a couch, the agent determines that a book should be placed on the coffee table, as shown in~\cref{fig:task}(c). 
For qualitative videos and additional examples, please refer the supplementary.

\textbf{Failure Modes}. The majority of \ESP evaluation failures for our CLIP-UNet eSP baseline come from the navigation and place planner.
In $53.5\%$ of cases, the navigation policy is unable to reach within $0.2m$ of the predicted \SP mask, as this requires precise navigation around clutter.
%
$31.0\%$ of the time, the place policy fails to execute fine-grained control to realize the highlighted \SP prediction.
In some instances, realizing \SP predictions is not feasible with the Stretch embodiment. For example, if a \SP mask indicates a placement at the center of a dining table, the robot may be unable to reach it due to the table's size and the maximum extension supported by the platform.
%
%
%
This highlights a key area for future work: learning \SP in an embodiment-aware manner to improve downstream performance.
In only $15.5\%$ of cases is the placement location predicted by the \SP mask is incorrect, such as when the \SP mask is placed on an incorrect receptacle. 
Please refer to the supplementary for videos of these failure modes.

%


\vspace{-5pt}
\section{Conclusion}
\vspace{-5pt}
\label{sec:conclusion}

We propose Semantic Placement (\SP), a novel task 
%
where, given an image and object type, a vision system must predict a binary mask highlighting semantically-meaningful regions in an image where that object could be placed.
Learning to predict the invisible is hard. We address this challenge by making visible objects invisible: we start with an image of an object in context and remove that object from the image via inpainting.
This automated data curation pipeline, leveraging inpainting and object detection models, enables us to supervise an end-to-end \SP prediction model, CLIP-UNet, using real-world data.
%
Our CLIP-UNet produces \SP predictions which generalize well to the real-world, are favored more by humans, and enable downstream robot execution.

\xhdr{Acknowledgements}.
We thank the PRIOR team at AI2 for feedback on the project idea. The Georgia Tech effort was supported in part by NSF, ONR YIP, and ARO PECASE. The views and conclusions contained herein are those of the authors and should not be interpreted as necessarily representing the official policies or endorsements, either expressed or implied, of the U.S. Government, or any sponsor.

\bibliographystyle{ieeetr}
\bibliography{strings,main}
\clearpage
\appendix

\section{Dataset Details}
\label{sec:data_supp}

\subsection{Data Generation}

In this section, we provide additional details about the data generation pipeline.

\textbf{(A) Query Image}. As mentioned in Sec.~\ref{sec:dataset}, the complete list of text queries to retrieve raw images from the LAION dataset includes \texttt{living room}, \texttt{bedroom}, and \texttt{kitchen}. 

\textbf{(B) Find Objects of Interest}. For each image, we use Detic~\cite{zhou2022detecting} and SAM~\cite{kirillov2023segment} to find segmentation masks of the $9$ object categories of interest.
First, we prompt Detic to find all instances of each of these $9$ categories within an image. If no object is detected, the image is discarded.
Next, for each detected object instance, we compute the center point ($center_x$, $center_y$) of its bounding box [$x_1$, $y_1$, $x_2$, $y_2$] and use this center point as a prompt for SAM~\cite{kirillov2023segment} to predict a segmentation mask.
Among the $3$ masks predicted by SAM~\cite{kirillov2023segment}, we choose the one with the highest confidence for downstream inpainting.

\begin{figure*}[h]
  \centering
    \includegraphics[width=0.95\textwidth]{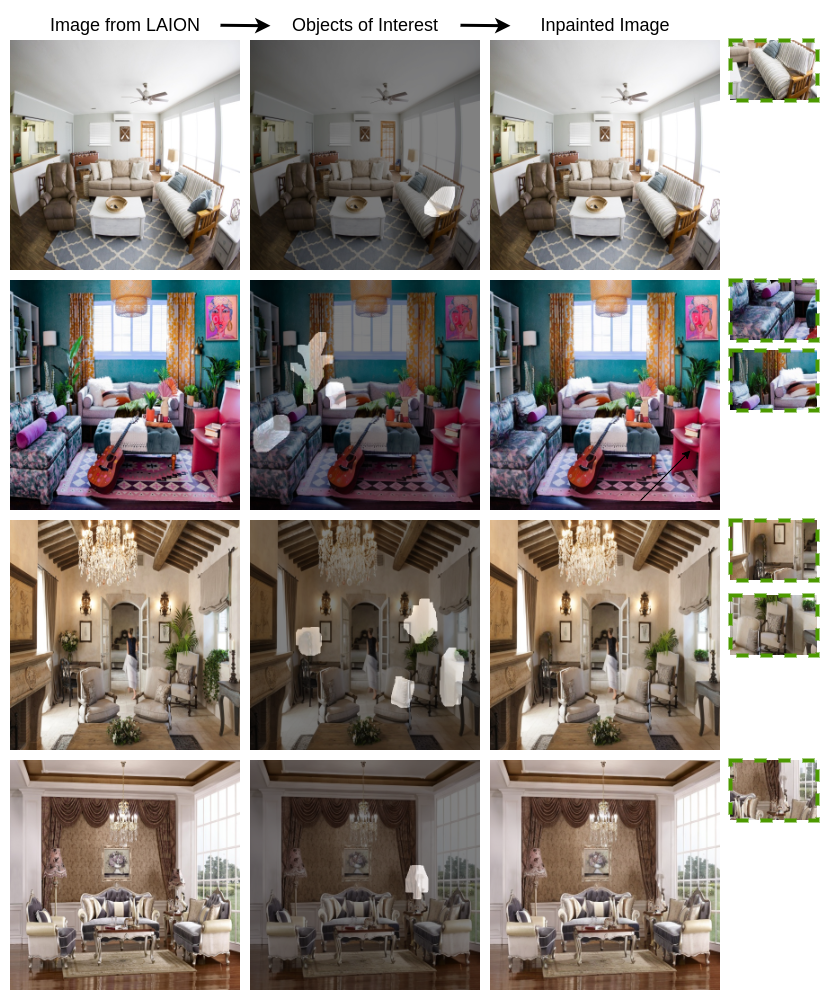}
    \vspace{5pt}
    \caption{Qualitative examples from \SP train data generated using our proposed automatic data generation pipeline.}
    \vspace{-1.5em}
    \label{fig:qualitative_data_example_1}
\end{figure*}
\begin{figure*}[h]
  \centering
    \includegraphics[width=0.95\textwidth]{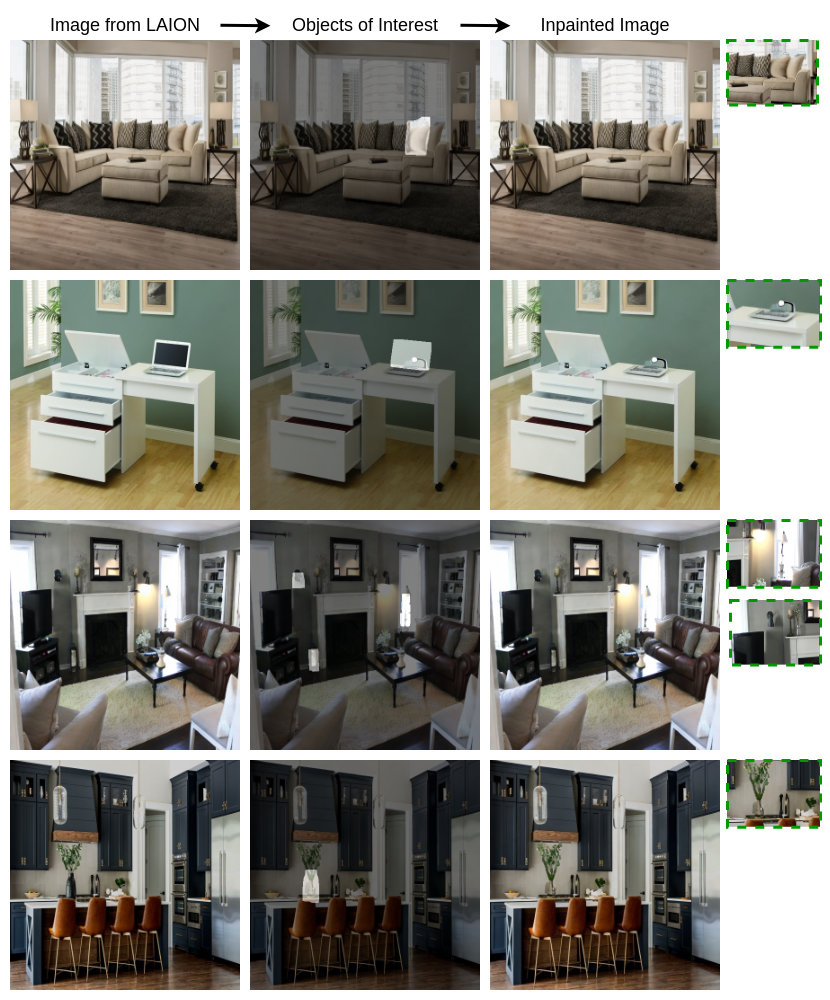}
    \vspace{5pt}
    \caption{Qualitative examples of inpainting failure during \SP data generation using our proposed automatic data generation pipeline.}
    \vspace{-1.5em}
    \label{fig:qualitative_data_example_2}
\end{figure*}

In~\cref{fig:qualitative_data_example_1} we visualize additional qualitative examples from the \SP training dataset generated using our automatic data generation pipeline.
Additionally, we also visualize examples of failures detected by our Detic filter and failed inpainting examples in ~\cref{fig:qualitative_data_example_2}.

\subsection{HSSD Image Dataset}
\label{sec:hssd_dataset}
To finetune our CLIP-UNet model for \SP mask prediction on a high-quality image dataset free from inpainting artifacts, we utilize the Habitat~\cite{savva_iccv19,andrew_neurips21} simulator along with the HSSD~\cite{khanna2023hssd} scene dataset.
HSSD is a synthetic indoor environment dataset comprising $211$ high-quality 3D scenes, containing $18{,}656$ models of real-world objects.
We generate the HSSD image dataset using the Habitat simulator, which allows us to manipulate scenes to render images with or without object, thereby avoiding any artifacts that models could exploit.
The training dataset consists of ${\sim}80k$ images generated using $135$ train scenes with
$8$ object categories.
Similarly, we create an evaluation dataset of ${\sim}18k$ images using $33$ val scenes 
with $8$ object categories.
Next, we describe the details of our image sampling process for different objects using the simulator.

\textbf{Image Sampling}.
To generate images from diverse viewpoints for each object instance, we first sample a set of candidate camera poses determined from polar coordinates $(r, \theta)$ relative to the object centroid, where $r \in \{0.5m, 1.0m, 1.5m, 2m\}$ and $\theta \in \{0^{\circ}, 10^{\circ} ,..., 360^{\circ} \}$.
We sample two types of viewpoints:

\begin{itemize}
    \item \textbf{Looking at Object}: For images looking at the objects of interest, we capture images with the camera's principal axis parallel to a ray from the camera's center to the object's centroid. We only keep the frames where the object of interest covers at least $5\%$ of the frame. This step ensures the inclusion of images where the target object and a valid placement is visible.
    \item \textbf{Random Viewpoints}: To add diversity, we also generate images from random viewpoints. Specifically, we run a frontier exploration~\cite{frontier} navigation agent in the environment to achieve ${\sim}90\%$ coverage. We then randomly sample $N$ images from the navigation trajectory, with $N = 250$ in our case, and add them to our dataset. We run this navigation agent $3$ times from random locations in each scene. We do not apply any frame coverage constraint during this phase to include images where no possible placement for an object exists. 
\end{itemize}

After determining all the viewpoints for each object instance in a scene, we programatically generate images with and without objects, target placement mask, and receptacle masks to add to our dataset.

\subsection{Real Evaluation Dataset}

For our experiments in~\cref{sec:experiments}, we use a real image dataset comprising $400$ images, collected from the LAION dataset~\cite{schuhmann2022laion} and $2$ real-world environments from~\cite{robothor,phone2proc}.
Specifically, this dataset includes $200$ images from the LAION dataset that were not seen during training, and an additional $200$ images from the real-world environments from~\cite{robothor,phone2proc}.

\section{Metric Details}
\label{sec:metrics_supp}

\subsection{Receptacle Priors}

To compute receptacle precision and recall metrics, we use the receptacles shown in~\cref{tab:obj_receptacle_mapping} for each object type from HSSD~\cite{khanna2023hssd} scenes.
To find the receptacle categories, we retrieve a list of receptacles that have an instance of the target object category placed on top, using metadata from the simulator.
It is important to note that since all \texttt{Trash Can} instances are usually found on the floor of an environment, there is no designated receptacle category for the \texttt{Trash Can} category.
Similarly, while some instances of the \texttt{Potted Plant} category are also found on the floor, we do not include \texttt{Floor} as a receptacle category.
This exclusion is due to the fact that the annotations for the \texttt{Floor} category cover the entire scene, making it challenging to quantify which part of the \texttt{Floor} annotation is a good or bad for object placement.


\begin{table}[t]
    \centering
    \resizebox{0.98\columnwidth}{!}{
        \begin{tabular}{@{}ll@{}}
            \toprule
            Object Category & Receptacles \\
            \midrule
            Cushion        & Couch, Bed, Sofa, Armchair \\
            Potted Plant & Coffee Table, Table, Chest of Drawers, Shelve, Kitchen Counter \\
            Book & Coffee Table, Table, Shelves, Couch, Sofa \\
            Vase & Coffee Table, Table, Chest of Drawers, Shelf, Kitchen Counter \\
            Alarm Clock & Bedside Table, Table, Chest of Drawers \\
            Laptop & Bed, Desk, Coffee Table, Table \\
            Table Lamp & Bedside Table, Chest of Drawers \\
            Toaster & Kitchen Counter \\
            Trash Can & - \\
            \bottomrule
            \end{tabular}
    }
    \vspace{5pt}
    \caption{Mapping of receptacles for each object category.}
    \label{tab:obj_receptacle_mapping}
\end{table}

\subsection{Human Evaluation}

To assess the performance of various methods on the Semantic Placemen (SP) task, we conduct a human evaluation study using Amazon Mechanical Turk.
Specifically, we conduct a user preference experiment in which human annotators are asked to compare \SP mask predictions from $5$ models (baselines from~\cref{tab:main_results}) and rank them from most to least preferred.
We conduct two types of the user study: one with the real image dataset and another with images from the HSSD~\cite{khanna2023hssd} scene dataset used in our experiments.
For each study, we randomly select $400$ images from the evaluation split of the respective datasets.
Each Amazon Mechanical Turk worker is assigned $20$ images to evaluate preferences, and each worker is allowed to participate in the study only once.
We report percentage of times annotators rank each model's \SP predictions as the best (\ie ranked above all other \SP predictions) in~\cref{tab:main_results} of the main paper.


\section{Baseline Details}
\label{sec:baseline_supp}

\textbf{Prior + Detector}.
For this baseline we leverage common-sense priors available in LLMs to find target receptacles for a particular object and use a open-vocabulary detector, Detic~\cite{zhou2022detecting}, to localize the receptacle in the image.
For each of the $9$ object categories in the dataset we prompt an LLM for common receptacle categories on which each object is found in indoor environment, shown in ~\cref{tab:obj_receptacle_mapping_baseline}.
Next, during evaluation we use object detector to localize the segmentation mask of all valid receptacles for a object category in an image.

\begin{table}[t]
    \centering
    \resizebox{0.98\columnwidth}{!}{
        \begin{tabular}{@{}ll@{}}
            \toprule
            Object Category & Receptacles \\
            \midrule
            Cushion        & Couch, Bed, Sofa, Armchair, Bench \\
            Potted Plant & Window Sill, Table, Chest of Drawers, Shelve, Balcony \\
            Book & Coffee Table, Table, Bookshelf, Desk, Nightstand, Bed \\
            Vase & Coffee Table, Table, Shelf, Mantle, Window Sill \\
            Alarm Clock & Bedside Table, Nightstand, Desk, Shelf \\
            Laptop & Desk, Table, Workstation \\
            Table Lamp & Desk, Nightstand, End Table, Shelf \\
            Toaster & Kitchen Counter, Shelf, Pantry \\
            Trash Can & Kitchen, Bathroom, Bedroom, Office \\
            \bottomrule
            \end{tabular}
    }
    \vspace{5pt}
    \caption{\textbf{Prior + Detector Baseline.} Mapping of receptacles from a LLM for each object category.}
    \label{tab:obj_receptacle_mapping_baseline}
\end{table}

\textbf{LLaVA}. VLMs like LLaVa~\cite{liu2023improvedllava} connect vision encoders to LLMs, enabling general purpose vision-and-language understanding.
To evaluate LLaVA on the \SP task, given an input image, we prompt it to output normalized bounding box coordinates for localizing a placement area.
The prompt we use is as follows: \\

``\texttt{You are a smart assistive robot tasked with cleaning this house. Localize the area in image as a bounding box in normalized coordinates to place the <object\_category>}''. \\

\begin{figure*}[h]
  \centering
    \includegraphics[width=0.95\textwidth]{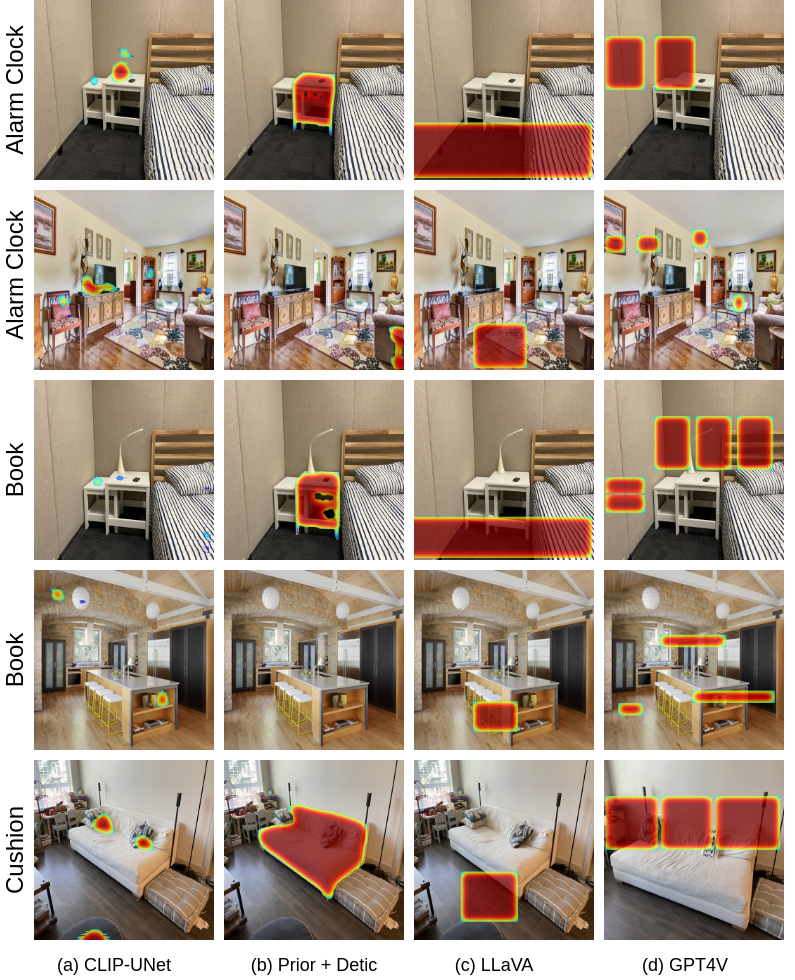}
    \vspace{5pt}
    \caption{Qualitative examples of \SP masks predicted by all the baselines on \SP Real val dataset}
    \label{fig:qualitative_example_1}
\end{figure*}

\begin{figure*}[h]
  \centering
    \includegraphics[width=0.95\textwidth]{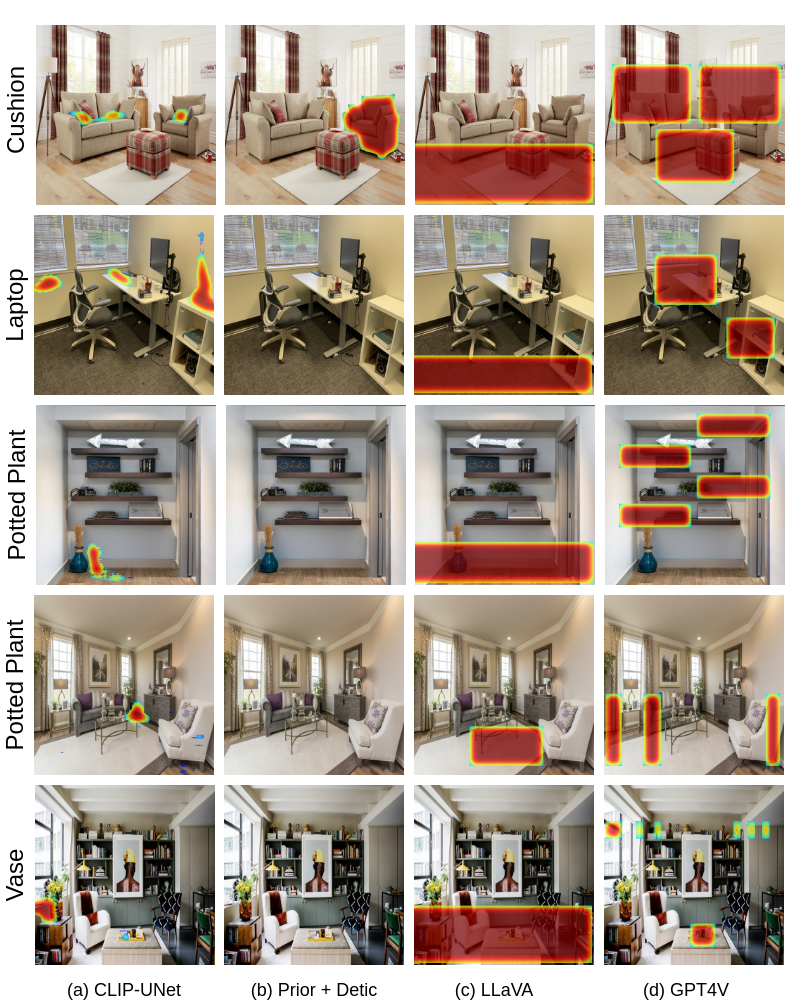}
    \vspace{5pt}
    \caption{Qualitative examples of \SP masks predicted by all the baselines on \SP Real val dataset}
    \label{fig:qualitative_example_2}
\end{figure*}

Subsequently, we convert the predicted normalized bounding box into a binary segmentation mask, which is then used as the \SP mask predictions for downstream applications.
Refer~\cref{fig:qualitative_example_1} and ~\cref{fig:qualitative_example_2} for qualitative examples.

\textbf{GPT4V~\cite{openai2023gpt4}}. Similar to LLaVA~\cite{liu2023improvedllava}, GPT4V is a multimodal LLM renowned for its vision-and-language understanding capabilities.
To evaluate GPT4V for the \SP task, we feed it an input image and prompt it to output normalized bounding box coordinates. These coordinates are then localized to a placement area and converted into a binary segmentation mask for use as \SP mask predictions.
We use the following prompt:

``\texttt{Here is an image of an indoor living environment.
We would like to determine all places in the image where one could potentially place an object of type <object\_type> so that environment remains tidy.
For example, you should not place a blender on the floor as blenders are not typically found on the floor.}
\\

\texttt{Please respond, in text, with a list of bounding box coordinates of potential locations. These bounding box coordinates should be of the form} \\

\texttt{[min x, min y, max x, max y]} \\

\texttt{where x and y are 0-1 valued and correspond to the fraction of the image along the width and height of the image with the top left of the image as the origin. 
Each set of coordinates should be on a new line. If there are no locations in the image where a <object\_type> could be placed, respond only with `NONE'.
Respond ONLY with these coordinates or NONE, do not include any other text in your response.}''\\

Subsequently, the predicted normalized bounding boxes are converted to binary segmentation masks as \SP mask predictions for downstream evaluation.
Refer~\cref{fig:qualitative_example_1} and ~\cref{fig:qualitative_example_2} for qualitative examples.

\section{Qualitative Results}
\label{sec:results_supp}

In~\cref{fig:qualitative_example_1} and ~\cref{fig:qualitative_example_2}, we visualize qualitative examples from the CLIP-UNet, Prior + Detector (Detic), LLaVA, and GPT4V baselines. These examples are images in the \SP real evaluation split, which were used for human evaluation.

\begin{figure*}[t]
  \centering
    \includegraphics[width=0.95\linewidth]{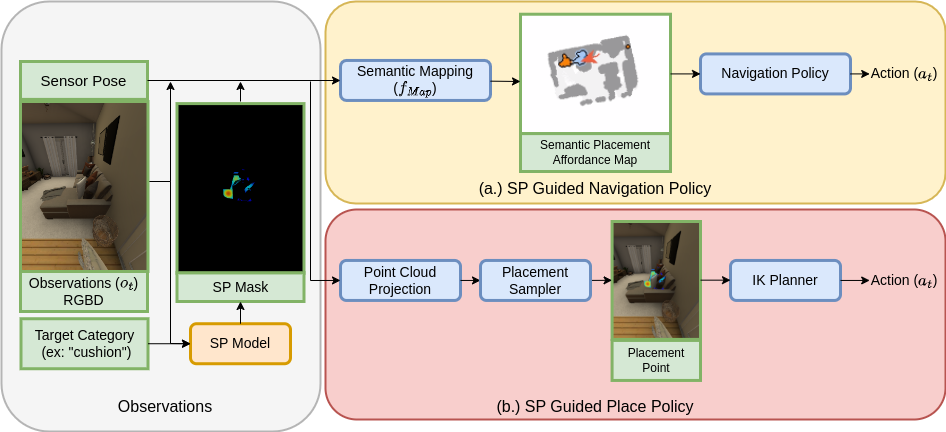}
    \vspace{5pt}
    \caption{\textbf{Embodied Evaluation Pipeline.} We build a two-stage modular policy consisting of: 1.) SP Guided Navigation Policy: Uses frontier exploration and semantic placement affordance 2D map to navigate to placement area, 2.) SP Guided Place Policy: Uses predicted \SP mask, projects it onto a pointcloud to sample placement point and uses IK planner to place the object.}
    \label{fig:embodied-eval}
\end{figure*}

\section{Embodied Evaluation Setup}
\label{sec:embodied_eval_setup}

In this section, we detail the Embodied Semantic Placement Policy used in~\cref{sec:embodied_eval} for evaluating the \ESP task of building a tidying robot.
%
%
Our experiments employ Hello Robot's Stretch robot~\cite{Kemp2022StretchRobot} with the full action space as defined in~\cite{yenamandra2023homerobot}. 
Specifically, the observation space, shown in the Fig.~\ref{fig:embodied-eval} Observations, includes RGB+Depth images from the robot's head camera, the camera pose, the robot's joint and gripper states, and the robot's pose relative to the starting pose of an episode.
The robot's action space comprises discrete navigation actions: \moveforward ($0.25m$),
\turnleft ($30^{\circ}$), \turnright ($30^{\circ}$), \lookup ($30^{\circ}$),
and \lookdown ($30^{\circ}$).
For manipulation, we use a continuous action space for fine-grained control of the gripper, arm extension and arm lift. 
The head tilt, pan and gripper's yaw, roll and
pitch can be changed by a maximum of $0.02-0.1$ radians in a single step, while the arm's extension and lift can be
changed by up to $2-10cm$ per step.
To perform the task with only the robot's observations and \SP mask predictions from the CLIP-UNet at each frame, we build a two-stage modular policy, comprising ``navigation'' and ``place'' policies, illustrated in Fig~\ref{fig:embodied-eval} (a.) \SP Guided Navigation Policy and (b.) \SP Guided Place Policy, respectively. The details for both policies are as follows.

\textbf{SP Guided Navigation Policy}.
Building upon the navigation policy from~\cite{chaplot_neurips20}, we replace the semantic map module with our semantic placement (\SP) map module.
To construct the \SP affordance map, we predict the \SP mask using egocentric observations at each timestep. This mask is then backprojected into a point cloud using
preceived depth. We bin the point cloud into a 3D \SP voxel map and sum it over height to derive the 2D \SP map.
Similar to~\cite{chaplot_neurips20}, our navigation policy employs frontier exploration~\cite{frontier}, using the 2D \SP map. 
We first build a \SP map by running the policy with the goal of maximizing coverage of the environment for $250$ steps. On average, we achieve about $60\%$ coverage of an environment within these $250$ steps.
Subsequently, the agent uses the \SP map to navigate towards the \SP mask instance that occupies the largest area on the 2D map. 

\textbf{SP Guided Place Policy}.
We build upon heuristic place policy from~\cite{yenamandra2023homerobot}. This policy assumes that the robot is within interactable distance (within $0.2m$) of the target receptacle where the object is to be placed.
First, the agent takes a panoramic turn until a valid \SP prediction is found (\ie not on the floor). This involves projecting the depth and \SP prediction onto a point cloud, transforming it into the agent's base coordinates, and applying a height filter.
Once a valid \SP prediction is identified, we estimate a placement point at the center of the largest slab (point cloud) for object placement on a flat surface.
To identify the largest flat surface slab, we score each point based on the number of surrounding points in the X/Y plane (with Z being up) within a $3cm$ height threshold, similar to~\cite{yenamandra2023homerobot}.
After determining the placement point, we rotate the robot to facing the point. This is required because the Stretch robot's arm is not aligned with the camera by default.
If the robot is at least $38.5cm$ away from the placement point, we move the robot forward, and re-estimate the placement point as described in~\cite{yenamandra2023homerobot}.
Finally, when the robot is sufficiently close, we use inverse kinematics to compute a sequence of actions to move the arm $15cm$ above the sampled voxel (to avoid collisions) to place (or drop) the object.

\section{Failure Modes}

In this section, we describe various failure modes of our CLIP-UNet model observed during its evaluation on the \SP task and in the downstream embodied evaluation of the \ESP task.

\subsection{Semantic Placement}

\begin{figure*}[h]
  \centering
    \includegraphics[width=0.95\textwidth]{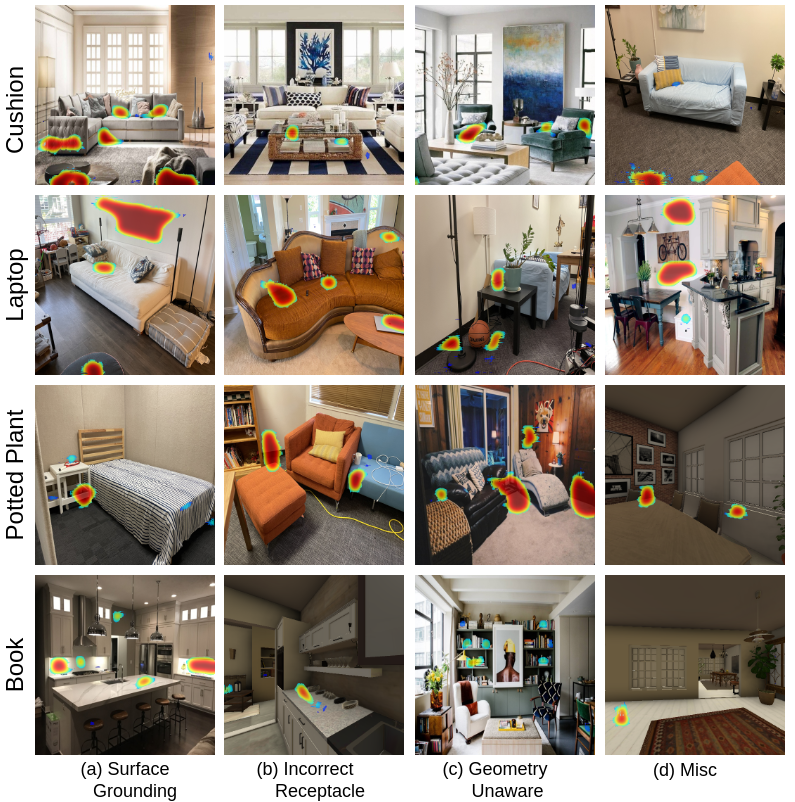}
    \vspace{5pt}
    \caption{Qualitative examples of failure modes of \SP mask prediction by our approach}
    \label{fig:failures_1}
\end{figure*}

Refer to~\cref{fig:failures_1} for examples of failure modes in \SP mask predictions by our CLIP-UNet model.
The common failure modes include:

\textbf{Surface Grounding}. Predictions that are not properly grounded to a surface of the receptacle in the image.

\textbf{Incorrect Receptacle}. Predictions with a $0$ Intersection over Prediction (IoP), indicating no overlap with any of the visible receptacles in the image.

\textbf{Geometry Unaware}. Our method, by design, is not capable of predicting \SP masks that are object shape aware (as our model's only knowledge about the object is the object's category). Consequently, we sometimes observe placements predicted by the model that are not geometry-aware, meaning the \SP masks highlight areas where there is insufficient space to place a new object.


\textbf{Misc}. This category contains all other failure cases, including predictions from the model that are noisy, placed on the floor/ceiling, or involve closed receptacles, etc.

\subsection{Embodied Semantic Placement}

The majority of \ESP evaluation failures come from the navigation and place planner, which include:

\textbf{Navigation Failure}. In $53.5\%$ of cases, the navigation policy fails to reach within $0.2m$ of the predicted \SP mask. This is often due to the requirement for precise navigation around clutter.

\textbf{Place Failure}. The place policy fails $31.0\%$ of the time to execute fine-grained control to realize the highlighted \SP prediction.
Occasionally, realizing \SP predictions is not feasible with the Stretch embodiment. For example, if a \SP mask indicates a placement at the center of a dining table, the robot might be unable to reach it due to the table's size and the maximum arm extension of the Stretch robot.
This highlights the need for future work in learning \SP in an embodiment-aware manner to improve downstream performance.

\textbf{Incorrect \SP Masks}. In $15.5\%$ of cases, the placement predicted by the \SP mask is incorrect, such as when the \SP mask is placed on an incorrect receptacle. 

Refer to the attached videos in the supplementary material for examples of these failure modes.

\section{Limitations}

Our approach is fundamentally constrained by the limitations of open-vocabulary object detectors, segmentation models, and inpainting models.
Since we employ these advanced ``foundation'' models off-the-shelf for automatic data generation, the quality of our generated data is heavily dependent on the performance of these models.
Moreover, the occasional poor performance of these models can introduce biases into the training dataset, which downstream models might exploit.
For example, false positive detections from open-vocabulary detectors (\eg a ceiling light detected as a lamp) may lead to biases in predicting \SP masks for lamps on the ceiling.
Similarly, imperfect inpainting models can produce artifacts like partially inpainted generations that bypass our detector-based validations, resulting in training data that may instill unrealistic biases in our model.
While finetuning on simulated data from HSSD can mitigate some of these biases, it might also introduce a domain gap for sim-to-real transfer.
Collecting high-quality real-world data for finetuning could help to alleviate this limitation.
Another challenge is that deploying the \SP prediction model zero-shot for applications like \ESP might yield \SP predictions that are not realizable given the robot's physical capabilities. 
A potential solution could involve finetuning the \SP model with the downstream task in an end-to-end manner. This aspect, however, remains as part of future work.



\begin{figure*}[h]
  \centering
    \includegraphics[width=0.95\textwidth]{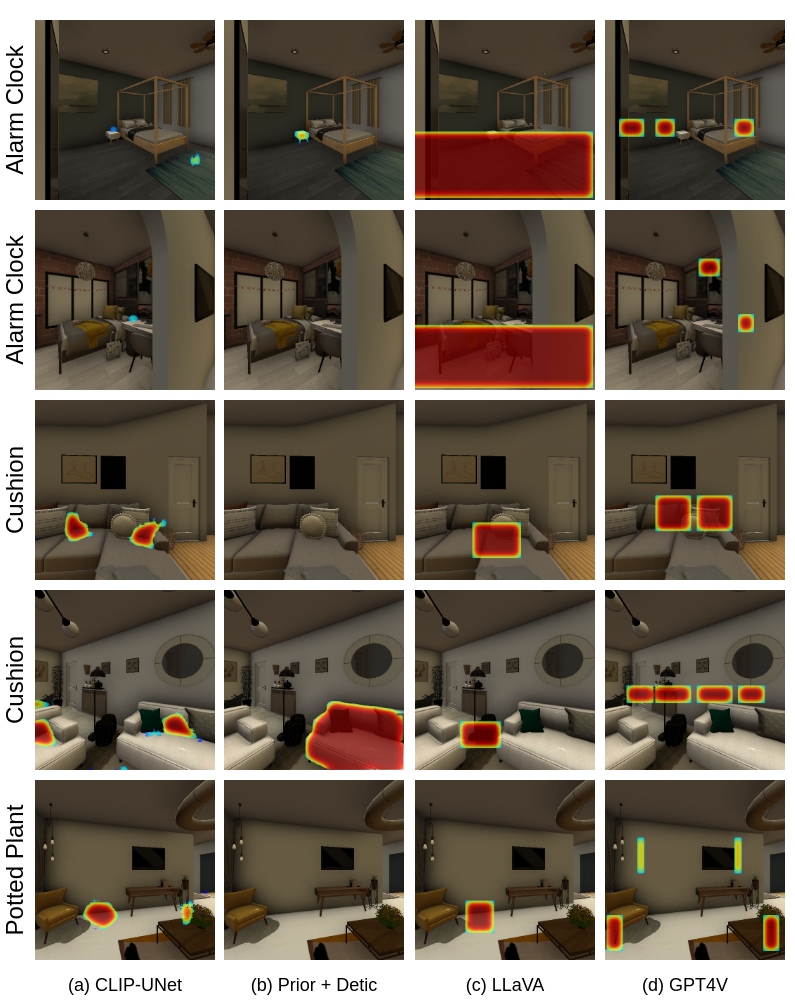}
    \vspace{5pt}
    \caption{Qualitative examples of \SP masks predicted by all the baselines on \SP HSSD val dataset}
    \label{fig:hssd_qualitative_example_1}
\end{figure*}

\begin{figure*}[h]
  \centering
    \includegraphics[width=0.95\textwidth]{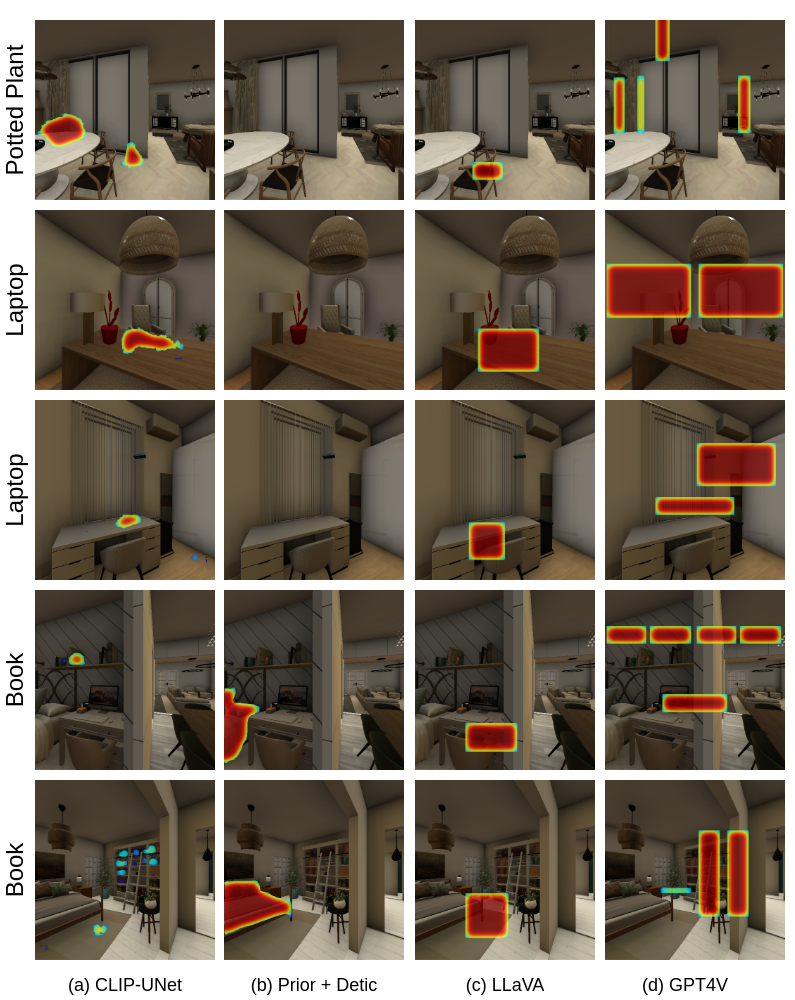}
    \vspace{5pt}
    \caption{Qualitative examples of \SP masks predicted by all the baselines on \SP HSSD val dataset}
    \label{fig:hssd_qualitative_example_2}
\end{figure*}

\end{document}